\documentclass[11pt]{article}

\usepackage[preprint]{acl}

\usepackage{times}
\usepackage{latexsym}

\usepackage[T1]{fontenc}
\usepackage[utf8]{inputenc}

\usepackage{microtype}

\usepackage{inconsolata}

\usepackage{graphicx}

\usepackage{subcaption}
\usepackage{booktabs} 
\usepackage{amsmath}
\usepackage{amssymb}

\usepackage{multirow}
\usepackage{colortbl}
\usepackage{tabularx}
\usepackage{float}
\usepackage{tcolorbox}
\definecolor{rowhighlight}{HTML}{FFF9D9}
\definecolor{subgroupcolor}{HTML}{F0F0F0}

\newcommand{\tbd}[1]{\textcolor{gray}{--}}

\title{Neural Procedural Memory: Empowering LLM Agents with Implicit Activation Steering}

\author{
 \textbf{Chengfeng Zhao\textsuperscript{1,2}},
 \textbf{Yuqiao Tan\textsuperscript{1,2}},
 \textbf{Shizhu He\textsuperscript{1,2}},
 \textbf{Yequan Wang\textsuperscript{3}},
 \textbf{Jun Zhao\textsuperscript{1,2}},
 \textbf{Kang Liu\textsuperscript{1,2}\thanks{Corresponding Author}},
\\
\\
 \textsuperscript{1}Institute of Automation, CAS
 \textsuperscript{2}University of Chinese Academy of Sciences
\\
 \textsuperscript{3}Beijing Academy of Artificial Intelligence
\\
\texttt{\{zhaochengfeng2024, tanyuqiao2025\}@ia.ac.cn}, \texttt{\{shizhu.he, kliu, jzhao\}@nlpr.ia.ac.cn} \\
\texttt{tshwangyequan@gmail.com} \\
}

\begin{document}
\maketitle
\begin{abstract}
While Large Language Models (LLMs) excel as static solvers, transforming them into autonomous agents remains challenging. 
This transition requires continuous environmental interaction, yet current agents lack the necessary persistent procedural memory.
Existing approaches predominantly employ Retrieval-Augmented Generation (RAG) to inject explicit textual guidelines into model contexts.
However, relying solely on symbolic instructions can introduce a text-action disconnect, frequently failing to activate the internal representations necessary for correct task execution.
To address this, the paper introduces \textbf{Neural Procedural Memory (NPM)}, a training-free framework that represents agent memory through implicit activation steering rather than explicit instructions.
By distilling procedural skills from historical contrastive experiences into steering vectors in the activation space, NPM directly activates the task-relevant neural mechanisms to guide task execution.
Evaluations across four agent benchmarks show that NPM performs comparably to baselines using explicit textual instructions. Furthermore, the results show that combining implicit steering with explicit workflows provides complementary advantages, leading to more robust task execution. 
Representational analyses indicate that these steering vectors encode consistent task logic, forming organized structures within the activation space. These findings suggest that implicit activation steering provides a promising approach for managing agent memory.
\end{abstract}

\section{Introduction}
\begin{figure}[t!]
    \centering
    \includegraphics[width=\columnwidth]{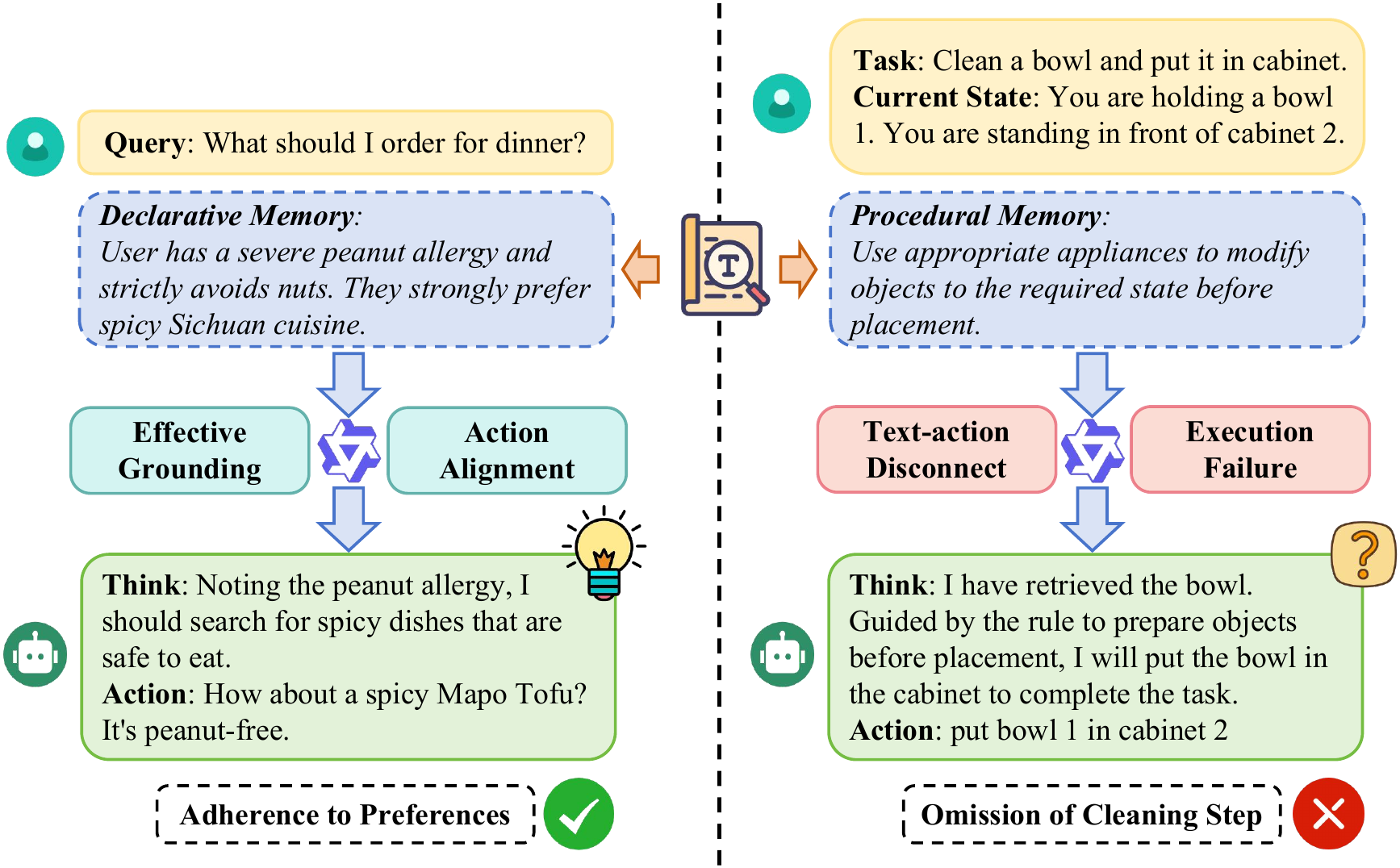}
    \vspace{-18pt}
    \caption{\textbf{Comparison between Declarative and Procedural Memory in LLM Agents.} \textbf{Left}: Declarative memory successfully grounds reasoning in static factual knowledge, enabling the agent to strictly adhere to explicit user constraints. \textbf{Right}: Textual procedural memory can introduce a text-action disconnect where the agent struggles to align the retrieved workflow onto the execution trajectory and omits intermediate steps.}
    \label{fig:intro}
    \vspace{-10pt}
\end{figure}

As Large Language Models (LLMs) evolve from stateless reasoners to the central controllers of autonomous agents, their operational scope has expanded to open-ended, dynamic environments---ranging from web navigation to embodied interaction~\cite{wang2023voyager,wang2024survey,liu2025advanceschallengesfoundationagents}. 
Succeeding in these complex settings requires agents to leverage persistent memory to transform past experiences into reusable skills, making memory a critical component of long-term autonomy~\cite{GenerativeAgents, MemoryBank, omidi2025memoryaugmentedtransformerssystematicreview, hu2025memoryageaiagents}.

Agent memory functionally consists of declarative memory for descriptive facts~\cite{wang2024aipersonalifelongpersonalization,gutierrez2024hipporag,Mem0} and procedural memory for executing action sequences~\cite{zheng2023synapse,ExpeL,yu2025browseragentbuildingwebagents}. While declarative recall is straightforward, effectively representing and utilizing procedural memory remains a bottleneck~\cite{han2025legomemmodularproceduralmemory,10.1145/3748302}. Traditional methods typically embed procedural knowledge offline into model parameters via supervised fine-tuning~\cite{shao-etal-2023-character,zhang2025agentlearningearlyexperience} or reinforcement learning~\cite{Room,yao2024retroformerretrospectivelargelanguage}. However, these parameter-updating approaches incur high computational overhead and struggle to adapt in real time to dynamic environments~\cite{Zhai2023InvestigatingTC,10763668, 11151751}.

By contrast, recent studies have focused on the second paradigm: online external knowledge retrieval, most notably Retrieval-Augmented Generation (RAG).
Although successful with declarative memory, such as fact retrieval~\cite{MemoryBank, MemGPT, Mem0}, this solution encounters limitations when applied directly to procedural memory.
Unlike static facts, procedural memory is inherently implicit and abstract. Such knowledge is often ineffable: compressing complex reasoning patterns into discrete natural language tokens inevitably leads to information loss~\cite{mahowald2024dissociating}. 
Relying exclusively on explicit textual descriptions to transfer procedural skills can lead to a text-action disconnect. Agents might comprehend retrieved instructions yet fail to strictly act upon them~\cite{wen2024benchmarkingcomplexinstructionfollowingmultiple,sun2024instructionfollowingevaluatinginferential,geng2025controlillusionfailureinstruction}. 
As shown in Figure~\ref{fig:intro}, while declarative memory successfully grounds static constraints, conveying procedural workflows through text fails to translate into sequential actions, leading agents to omit critical intermediate steps during execution.
When relying solely on texts as a reasoning medium, symbolic descriptions of actions can sometimes fail to consistently elicit the targeted behavioral trajectories required for task completion.

Therefore, it is intuitive that procedural memory could benefit from mechanisms more closely aligned with intrinsic behavioral modulation rather than relying solely on external text. Such a perspective is corroborated by cognitive neuroscience~\cite{SQUIRE.1992.4.3.232, SQUIRE2004171}, which posits that procedural memory is non-verbalizable and manifests through the modulation of neural activity rather than explicit declarative recall. 

Following this insight, we propose \textbf{Neural Procedural Memory (NPM)}, a training-free framework that introduces an alternative paradigm of agent memory based on implicit activation steering. Instead of retrieving text for the agent to read, NPM identifies relevant historical tasks and extracts procedural signals from contrastive experiences into steering vectors within the activation space.
Specifically, we employ a dual-granularity strategy to address the challenge of agents failing to obtain successful trajectories in complex tasks: inter-trajectory contrast aligns successful and failed trajectories, while intra-trajectory contrast exploits step-level differences within individual failed attempts.
During inference, the steering vectors are dynamically synthesized and injected into the residual stream, directly modulating the agent's reasoning process and action selection without expanding the context window or updating parameters.

We evaluate NPM across four agent benchmarks including ALFWorld~\cite{ALFWorld}, WebShop~\cite{WebShop}, ScienceWorld~\cite{wang2022scienceworld}, and BabyAI~\cite{chevalier-boisvert2018babyai}. Results demonstrate that implicit activation steering achieves performance comparable to explicit textual memory baselines and exhibits complementary synergy when combined with explicit textual workflows. While textual memory provides high-level symbolic guidance, implicit steering reinforces procedural adherence directly within the neural representation.
We further dissect the extracted vectors to reveal semantically meaningful behavioral primitives, providing an interpretable view of how procedural knowledge is represented in activation space. These findings suggest implicit activation steering as an effective path for representing procedural memory in LLM agents.

Our contributions are summarized as follows:
\begin{itemize}
    \vspace{-6pt}
    \item This paper proposes Neural Procedural Memory (NPM), a training-free framework that integrates implicit activation steering into agent memory, enabling effective intrinsic behavioral modulation.
    \vspace{-6pt}
    \item This paper introduces a dual-granularity extraction method using both inter-trajectory and intra-trajectory contrasts, offering a flexible intervention mechanism and mitigating the reliance on successful demonstrations.
    \vspace{-6pt}
    \item Extensive experiments demonstrate that activation steering serves as an effective carrier for agentic skills, yielding interpretable vectors that correlate with specific task behaviors.
\end{itemize}

\begin{figure*}[t!]
    \centering
    \includegraphics[width=0.9\textwidth]{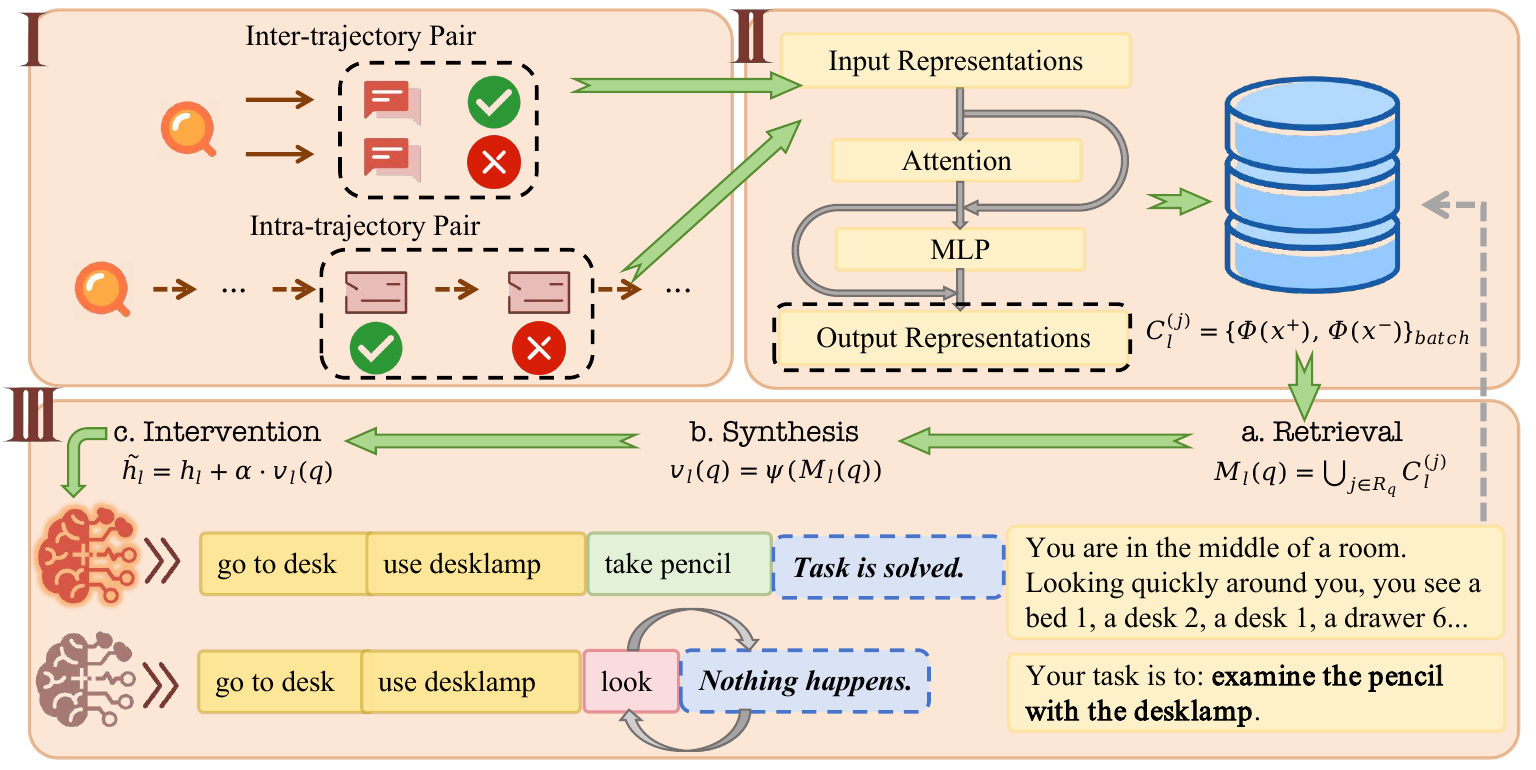}
    \vskip -8pt
    \caption{\textbf{Overview of the Neural Procedural Memory (NPM) framework.}
    (1) \textbf{Contrastive Experience Construction:} Formulating dual-granularity (inter- and intra-trajectory) contrastive pairs from historical interactions. (2) \textbf{Procedural Memory Extraction:} Extracting continuous representations from these pairs to construct a historical memory repository. (3) \textbf{Inference-Time Intervention:} Retrieving relevant experiences to dynamically synthesize a steering vector, which is injected into the residual stream to guide agent reasoning.}
    \vspace{-15pt}
    \label{fig:method_overview}
\end{figure*}

\vspace{-8pt}
\section{Related Work}
\vspace{-4pt}

\paragraph{Functions of Agent Memory}
Agent memory is typically categorized into factual (declarative) and experiential (procedural) components~\cite{hu2025memoryageaiagents}. While Retrieval-Augmented Generation (RAG) effectively maintains declarative facts~\cite{hu2025memoryageaiagents,GenerativeAgents}, it faces limitations when applied to procedural memory. 
Existing approaches typically approximate procedural memory by injecting textual guidelines or code snippets into the context window~\cite{shinn2023reflexion,wang2023voyager}. 
However, this explicit approach suffers from a text-action disconnect: natural language serves as an inefficient proxy for encoding internal neural representations~\cite{mahowald2024dissociating}.
Moreover, the reliance on lengthy text instructions increases computational cost and degrades performance in long-horizon tasks~\cite{geng2025controlillusionfailureinstruction,sinha2025illusiondiminishingreturnsmeasuring}. 
NPM shifts from explicit verbalization to implicit activation modulation, eliminating the need for in-context instructions.

\vspace{-8pt}
\paragraph{Forms of Agent Memory}
Current approaches explore various representational modalities. 
Token-level methods (e.g., RAG) offer flexibility but are bounded by context window limits and inference latency.
Conversely, parametric approaches encode knowledge within model weights permanently~\cite{zhang2025agentlearningearlyexperience,wang2025selfupdatable}, offering efficiency but lacking the flexibility required for rapid adaptation or unlearning without retraining.
Latent memory offers a middle ground by storing information as continuous vectors. However, existing works primarily focus on context compression~\cite{AutoCompressor,MEMORYLLM} or transient working memory~\cite{MemGen,VisMem}.
For instance, MemGen~\cite{MemGen} synthesizes latent tokens during reasoning but requires auxiliary encoders and extensive training.
While NPM shares the underlying concept of latent memory, it views memory storage as activation steering, enabling representation-level modulation of the generation path without additional token overhead or parameter updates.

\vspace{-8pt}
\paragraph{Representation Engineering and Activation Steering.}
Our framework builds upon Representation Engineering (RepE) that monitors and manipulates high-level cognitive states in the activation space~\cite{zou2023transparency}. Prior studies~\cite{rimsky-etal-2024-steering,lee2025programming} have demonstrated the efficacy of steering models toward coarse-grained, static attributes such as honesty or harmlessness. However, existing RepE work is predominantly limited to single-turn global concepts. NPM extends these techniques to dynamic, task-specific agentic trajectories. By introducing a retrieval-augmented framework where steering vectors are synthesized from contrastive historical experiences, we enable agents to adaptively recall and apply procedural skills specific to the task context.

% \vspace{-15pt}
\section{Neural Procedural Memory}
\vspace{-5pt}
We introduce \textbf{Neural Procedural Memory (NPM)}, a training-free framework designed to internalize agentic skills directly into the model's activation space.
The framework shown in Figure~\ref{fig:method_overview} consists of three phases: (1) \textbf{Contrastive Experience Construction} (\S\ref{sec: method_1}), which isolates effective reasoning patterns from historical failures; (2) \textbf{Steering Vector Extraction} (\S\ref{sec: method_2}), which distills discrete textual contrasts into continuous steering vectors; and (3) \textbf{Inference-Time Intervention} (\S\ref{sec: method_3}), where these vectors are dynamically synthesized and injected to modulate the agent's behavior.

\vspace{-5pt}
\subsection{Contrastive Experience Construction}
\label{sec: method_1}

Formally, given a task query $q$ and a task environment $\mathcal{E}$, an agent generates a trajectory $\tau = \{s_1, a_1, \dots, s_K, a_K\}$, where $s_k$ denotes the state and $a_k$ denotes the action at step $k$. 
The objective is to construct a contrastive dataset $\mathcal{D}$ composed of paired reasoning segments $(x^+, x^-)$, where $x^+$ represents a desirable reasoning mode and $x^-$ represents a degenerate one. A dual-granularity strategy is employed to capture procedural signals at both the inter-trajectory and intra-trajectory levels.

\subsubsection{Inter-trajectory Contrast}
\vspace{-1pt}
For tasks where the agent has historically generated both successful and failed attempts, we aim to capture the global behavioral shift. 
Let $\mathcal{T}^+$ and $\mathcal{T}^-$ denote the sets of successful and failed trajectories for a specific task, respectively. A contrastive pair is formed by aligning a failed attempt with a successful one:

\vspace{-15pt}
\begin{equation}
    \mathcal{P}_{\text{inter}} = \big\{ (\tau^+, \tau^-) \mid \tau^+ \in \mathcal{T}^+, \tau^- \in \mathcal{T}^- \big\}.
\end{equation}

\vspace{-5pt}
This contrast directionally distinguishes successful trajectories from failed attempts.

\vspace{-5pt}
\subsubsection{Intra-trajectory Contrast}
\vspace{-1pt}
In many complex tasks, the agent frequently fails to obtain successful trajectories, resulting in a sparse distribution of positive samples. To enable learning from failure, we exploit step-level differences within a single failed trajectory.
A \textbf{Degenerate Step} set $S_{\text{deg}} \subset \tau$ is defined to contain actions that exhibit irrational behaviors or violate environmental constraints, identified via the following criteria:
\begin{itemize}
    \vspace{-5pt}
    \item \textbf{Redundancy:} Consecutive repetition of an identical command, indicating a heuristic loop or failure to progress without state change.
    \vspace{-5pt}
    \item \textbf{Invalidity:} Actions that trigger environment-specific error feedback, such as format violations or inadmissible moves.

\end{itemize}
Conversely, an \textbf{Effective} step set $S_{\text{eff}} = \tau \setminus S_{\text{deg}}$ is defined to contain steps validly advancing the state.

Instead of pairing individual steps, which serves as a noisy signal, we aim to contrast the collective effective behavior against the degenerate one within the trajectory. The contrastive pair is constructed by grouping the steps into two sets:
\vspace{-5pt}
\begin{equation}
    \mathcal{P}_{\text{intra}} = \big\{ (S_{\text{eff}}, S_{\text{deg}}) \mid S_{\text{deg}} \neq \emptyset \big\}.
\end{equation}

\vspace{-6pt}
This intra-trajectory contrast isolates specific degenerate modes from the otherwise valid reasoning process. (Quantitative analysis of heuristic isolation rules is provided in Appendix~\ref{app:heuristic_rules}).

\vspace{-5pt}
\subsection{Procedural Memory Extraction}
\label{sec: method_2}

The core of NPM is to map the textual contrast defined in $\mathcal{D}$ into a geometric transformation in the continuous representation space. The goal is to derive a vector $\mathbf{v} \in \mathbb{R}^d$ representing the shift from a degenerate state $\mathbf{h}^-$ to a desirable one $\mathbf{h}^+$.
\vspace{-5pt}

\paragraph{Hidden State Representation.}
We define the representation function $\phi_l(\cdot)$ to extract a fixed-dimensional representation vector $\mathbf{h} \in \mathbb{R}^d$ depending on the contrast granularity. 
\begin{itemize}
\vspace{-8pt}
    \item \textit{Inter-trajectory:} For this comparison, we utilize the sequence of hidden states of the full trajectory $\tau$ and extract the last token to capture the accumulated context for each layer.
    \vspace{-20pt}
    \item \textit{Intra-trajectory:} Since reasoning steps span multiple tokens, we apply mean-pooling over the tokens to obtain a step representation:
    \vspace{-5pt}
    \begin{equation}
        \phi_l(S) = \frac{1}{|S|} \sum_{s \in S} \left( \frac{1}{|s|} \sum_{i=1}^{|s|} \mathbf{h}_{l,i}^{(s)} \right).
    \end{equation}
    
    \vspace{-10pt}
    Here, $|S|$ denotes the total number of steps in the set, $|s|$ denotes the length of an individual step $s$, and $\mathbf{h}_{l,i}^{(s)}$ corresponds to the activation of the $i$-th token within step $s$ at layer $l$.
\end{itemize}

\vspace{-10pt}
\paragraph{Memory Storage.}
For each historical task $Q_j$, we organize its extracted representations into a task-specific contrastive pair set $\mathcal{C}_l^{(j)} = \{ (\phi_l(x_i^+), \phi_l(x_i^-)) \}_{i=1}^{N_j}$, where $N_j$ denotes the total number of contrastive pairs collected for the task and $(x_i^+, x_i^-)$ represents the $i$-th pair corresponding to the granularities defined in \S\ref{sec: method_1}. These pre-computed sets serve as the raw procedural memory stored in our repository.

\vspace{-5pt}
\subsection{Inference-Time Intervention}
\vspace{-2pt}
\label{sec: method_3}
NPM applies procedural memory dynamically at inference time without modifying model weights. To ensure low latency, we pre-compute and store the hidden state representations defined in §\ref{sec: method_2} for all historical experiences. The inference process follows a Retrieval-Synthesis-Intervention pipeline:

\begin{enumerate}
\vspace{-8pt}
    \item \textbf{Contextual Retrieval:} Given a new test task $q$, we utilize a dense retriever to identify the top-$K$ most similar historical tasks $\mathcal{R}_q$.
    \vspace{-5pt}
    \item \textbf{Dynamic Vector Synthesis:} We fetch the stored representation sets for the retrieved tasks to form a collective memory pool: $\mathcal{M}_l(q) = \bigcup_{j \in \mathcal{R}_q} \mathcal{C}_l^{(j)}$. The synthesis strategy selects between inter- and intra-trajectory contrasts. A task-specific consensus direction $\mathbf{v}_l(q)$ is then derived from the collective memory pool via extraction function $\psi(\cdot)$:
    \vspace{-5pt}
    \begin{equation}
        \mathbf{v}_l(q) = \psi\big(\mathcal{M}_l(q)\big)
    \end{equation}
    \vspace{-25pt}
    \item \textbf{Activation Steering:} During autoregressive generation, we inject the synthesized vector into the residual stream at each time step $t$. For a target layer $l$, the intervened activation $\tilde{\mathbf{h}}_{l,t}$ is computed as:
    \vspace{-5pt}
    \begin{equation}
        \tilde{\mathbf{h}}_{l,t} = \mathbf{h}_{l,t} + \alpha \cdot \mathbf{v}_l(q)
    \end{equation}

    \vspace{-8pt}
    where $\mathbf{h}_{l,t}$ denotes the original hidden state of the current token, and $\alpha$ is a scalar parameter controlling the intervention strength.
\end{enumerate}
\vspace{-5pt}
By shifting the activation distribution, NPM implicitly nudges the agent's intuition away from known failure modes and towards effective reasoning paths. Detailed storage and latency analyses are provided in Appendix~\ref{app:system_efficiency}.

\vspace{-5pt}
\section{Experiments}
\subsection{Experimental Setup}

We evaluate the proposed framework using MiniCPM3-4B~\cite{hu2024minicpm} and Qwen3 models (4B and 8B)~\cite{qwen3} across four agent benchmarks: ALFWorld~\cite{ALFWorld}, WebShop~\cite{WebShop}, ScienceWorld~\cite{wang2022scienceworld}, and BabyAI~\cite{chevalier-boisvert2018babyai}. The evaluation relies on the success rate for ALFWorld and the average reward for the remaining environments. Detailed descriptions of each simulated environment and task configurations are deferred to Appendix~\ref{app:implementation}.

We compare \textbf{NPM} against baselines representing no memory, explicit textual memory, and implicit activation steering. The explicit baselines augment the model context, utilizing \textbf{Insights}~\cite{ExpeL} for global procedural guidelines or \textbf{Workflows}~\cite{AWM} for concrete execution patterns. The implicit steering baselines include \textbf{CAA}~\cite{panickssery2024CAA} and \textbf{Mass-Mean}~\cite{marks2024massmean}, applying fixed activation shifts computed across the entire dataset. Our approach differs from these static methods by dynamically synthesizing task-specific intervention vectors from contrastive experiences. We additionally evaluate a \textbf{Hybrid} configuration combining implicit steering with explicit workflows to examine whether the two representation modalities offer complementary advantages.

\vspace{-5pt}
\subsection{Main Results}
\begin{table*}[t!]
\centering
\caption{\textbf{The overall performance comparison.} We report the success rate (\%) for ALFWorld and the average reward for WebShop, ScienceWorld, and BabyAI. ``Avg'' denotes the average score across the four benchmarks.}
\vspace{-5pt}
\label{tab:main_results}
\resizebox{\textwidth}{!}{%
\begin{tabular}{ c l l c c c c c}
\toprule
\textbf{Model} & \textbf{Paradigm} & \textbf{Method} & \textbf{ALFWorld} & \textbf{WebShop} & \textbf{ScienceWorld} & \textbf{BabyAI} & \textbf{Avg} \\
\midrule
\multirow{7}{*}{\textbf{MiniCPM3-4B}}
& None       & No Memory    & 23.88 & 32.21 & 7.29 & 27.00 & 22.60 \\
\cmidrule(lr){2-8}
& \multirow{2}{*}{Explicit} & Insights & 28.36 & 47.96 & 10.0 & 27.78 & 28.53 \\
&                           & Workflows     & 32.84 & \textbf{54.97} & 10.5 & 32.39 & 32.68 \\
\cmidrule(lr){2-8}
& \multirow{3}{*}{Implicit} & CAA          & 23.88&	13.85&	6.29&	30.29&	18.58
 \\
&                           & Mass-Mean    & 26.87&	34.21&	8.02&	29.73&	24.71 \\
&                           & NPM (Ours)   &31.34&	43.53	&8.70	&31.91&	28.87 \\
\cmidrule(lr){2-8}
& Hybrid     & NPM + Workflows    & \textbf{38.81} &	51.21&	\textbf{10.97}&	\textbf{36.90}&	\textbf{34.47} \\
\midrule
\multirow{7}{*}{\textbf{Qwen3-4B}}
& None       & No Memory    & 30.60 & 44.81 & 18.40 & 18.75 & 28.14 \\
\cmidrule(lr){2-8}
& \multirow{2}{*}{Explicit} & Insights & 44.03	& 40.72 & 22.59 & 12.09 & 29.86 \\
&                           & Workflows  & 52.99	& 45.73 & \textbf{23.59} & 14.60 & 34.23 \\
\cmidrule(lr){2-8}
& \multirow{3}{*}{Implicit} & CAA         & 26.87&	25.52&	11.48&	3.48&	16.84
 \\
&                           & Mass-Mean   & 29.10&	39.07&	18.16&	17.82	&26.04 \\
&                           & NPM (Ours)   & 40.30&	\textbf{48.00}&	18.20&	19.04&	31.39 \\
\cmidrule(lr){2-8}
& Hybrid     & NPM + Workflows    & \textbf{61.19}&	47.84&	21.97&	\textbf{19.38}&	\textbf{37.60} \\
\midrule
\multirow{7}{*}{\textbf{Qwen3-8B}}
& None       & No Memory    & 39.55	& 46.25 & 24.98 & 11.74 & 30.63  \\
\cmidrule(lr){2-8}
& \multirow{2}{*}{Explicit} & Insights & 46.27	& 45.22 & 28.09 & 7.77 & 31.84 \\
&                           & Workflows     & 62.69 & 48.93 & 29.57 & 10.40 & 37.90 \\
\cmidrule(lr){2-8}
& \multirow{3}{*}{Implicit} & CAA          & 47.01&	20.66	&9.17&	14.31&	22.79
 \\
&                           & Mass-Mean    & 49.25&	44.26&	25.32&	8.71&	31.89 \\
&                           & NPM (Ours)   & 56.72&	48.24&	26.07&	14.26&	36.32 \\
\cmidrule(lr){2-8}
& Hybrid     & NPM + Workflows    & \textbf{66.42}&	\textbf{53.94}&	\textbf{31.89}&	\textbf{15.31}&	\textbf{41.89} \\

\bottomrule
\end{tabular}%
}
\vspace{-15pt}
\end{table*}

The experimental results are presented in Table~\ref{tab:main_results}.

\vspace{-5pt}
\paragraph{Effectiveness of Implicit Steering.}
NPM improves performance across most evaluated model configurations and environments compared to the base model without memory. For example, the average score for MiniCPM3-4B increases from 22.60 to 28.87 with implicit steering, and Qwen3-8B improves from 30.63 to 36.32. Furthermore, NPM outperforms static baselines such as CAA and Mass-Mean. These alternative methods struggle to capture multi-step procedural skills since they rely on fixed mean differences computed across the entire dataset. The performance difference indicates that retrieving contextually relevant historical pairs and adapting the intervention to the specific task provides a more accurate behavioral correction.

\vspace{-5pt}
\paragraph{Competitiveness with Explicit Memory.}
Implicit activation steering achieves results comparable to explicit textual paradigms. On average, NPM outperforms the Insights approach, demonstrating that direct intrinsic modulation can be more effective than broad textual principles. While the more structured Workflows baseline shows stronger performance by offering explicit step-by-step guidance, NPM remains competitive and outperforms in certain settings. On the WebShop using Qwen3-4B, NPM scores 48.00, exceeding the 45.73 achieved by explicit workflows. While textual workflows provide specific steps, translating these textual instructions into actions can introduce a text-action disconnect during long sequences (Detailed qualitative case study is provided in Appendix~\ref{app:case_study}). Furthermore, maintaining extensive textual memory consumes context window space and increases processing overhead. 
NPM offers an efficient alternative by guiding the model directly in the activation space, delivering competitive performance without occupying context window tokens.

\vspace{-5pt}
\paragraph{Complementary Synergy of Hybrid Memory.}
Integrating implicit steering with explicit workflows yields the highest overall performance, setting the top average scores across all three backbone models. Combining both memory forms on Qwen3-8B elevates the success rate on ALFWorld to 66.42 percent and reaches a peak score of 31.89 on ScienceWorld. This synergy indicates that explicit text and implicit representations operate as a complementary system. Textual workflows supply high-level symbolic planning steps, whereas NPM provides direct neural intervention to ensure adherence to these plans during long execution horizons. This persistent modulation helps prevent the agent from deviating from prompt-based instructions during extended interactions.

\vspace{-5pt}
\subsection{Dual-Granularity Contrasts}

\label{sec:ablation_dual_granularity}
\begin{figure}[htbp!]
    \centering
    \includegraphics[width=\columnwidth]{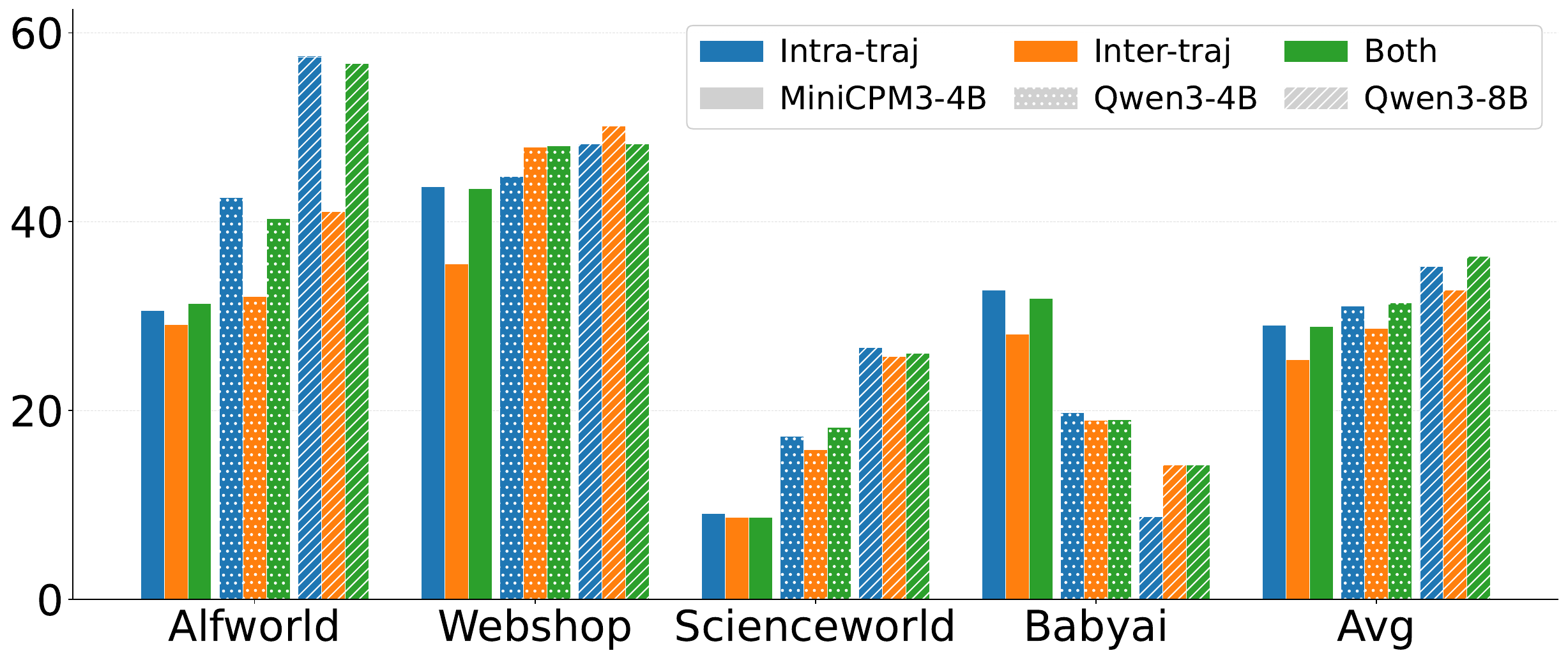}
    \vspace{-20pt}
    \caption{Performance comparison of different steering granularities across multiple environments and models.}
    \label{fig:dual_granularity}
    \vspace{-10pt}
\end{figure}

Tasks across different environments vary in complexity, therefore a single level of intervention may not always provide the appropriate correction. We evaluate the individual effects of intra-trajectory and inter-trajectory steering to understand this behavior. As shown in Figure~\ref{fig:dual_granularity}, relying exclusively on a single granularity results in performance variance across different benchmarks and model architectures. Intra-trajectory steering performs well in step-intensive environments such as ALFWorld by correcting local action errors. Inter-trajectory steering proves more effective when agents require macro-level planning guidance like WebShop.
To balance these effects, we dynamically select the appropriate intervention granularity based on the task context. 
This combination reduces performance variance and provides stable generalization across configurations.
Although the implemented selection strategy occasionally yields lower success rates than an oracle baseline in certain boundary cases, the empirical results confirm that the dual-granularity contrast provides a reliable mechanism for managing procedural interventions. The representational differences underlying these two steering granularities are further examined in Section~\ref{sec:ana}.

\vspace{-5pt}
\section{Analysis}
\label{sec:ana}
\vspace{-5pt}
\subsection{Representational Separability of Activations}
\label{sec:analysis_geometry}

\begin{figure}[t]
    \centering
    \begin{subfigure}[b]{0.48\columnwidth}
        \centering
        \includegraphics[width=\columnwidth]{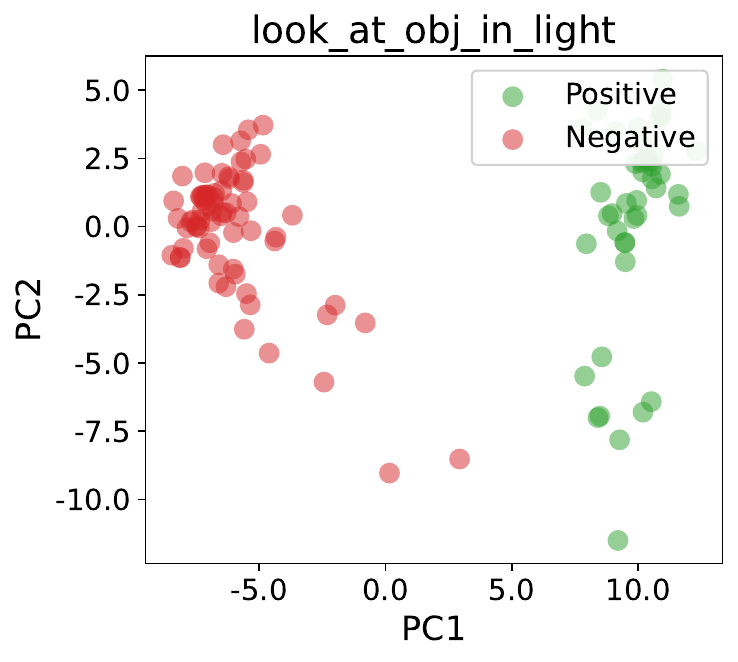}
        \caption{Inter}
        \label{fig:inter_a}
    \end{subfigure}
    \hfill
    \begin{subfigure}[b]{0.48\columnwidth}
        \centering
        \includegraphics[width=\columnwidth]{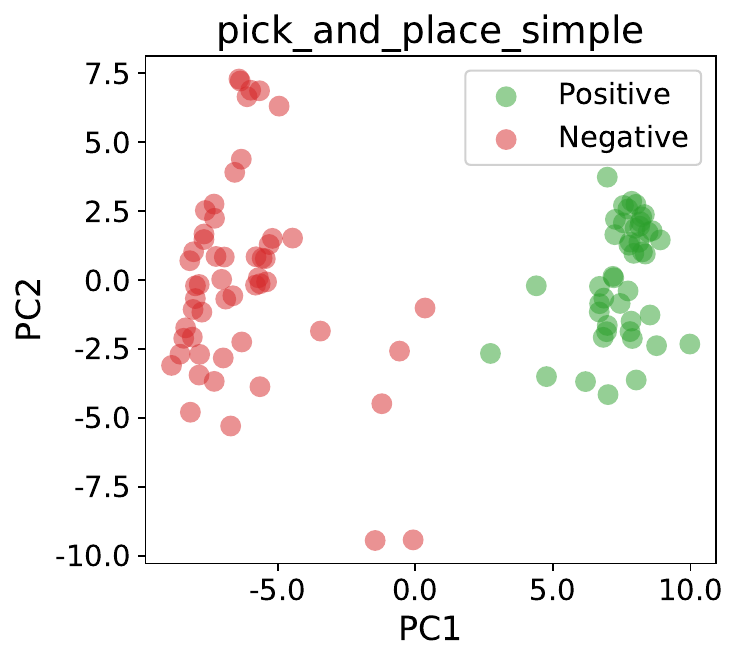}
        \caption{Inter}
        \label{fig:inter_b}
    \end{subfigure}

    \begin{subfigure}[b]{0.48\columnwidth}
        \centering
        \includegraphics[width=\columnwidth]{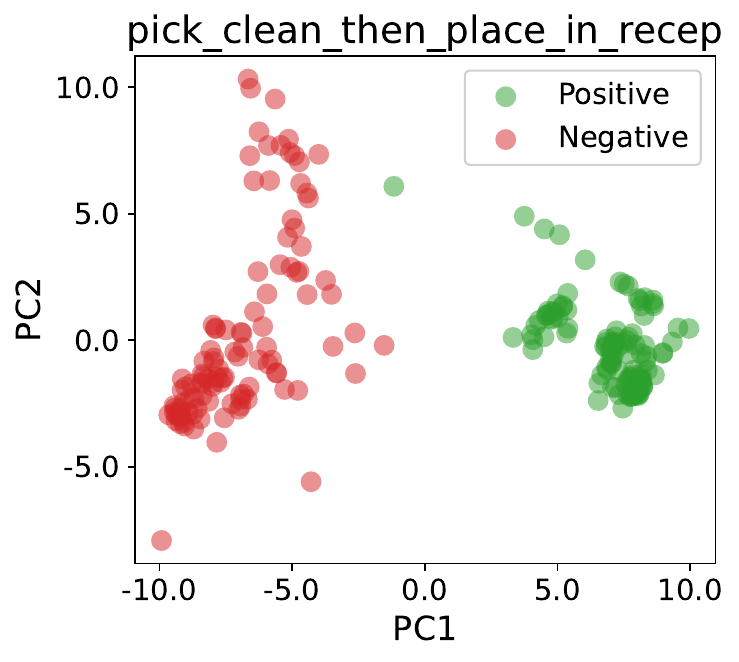}
        \caption{Intra}
        \label{fig:intra_a}
    \end{subfigure}
    \hfill
    \begin{subfigure}[b]{0.48\columnwidth}
        \centering
        \includegraphics[width=\columnwidth]{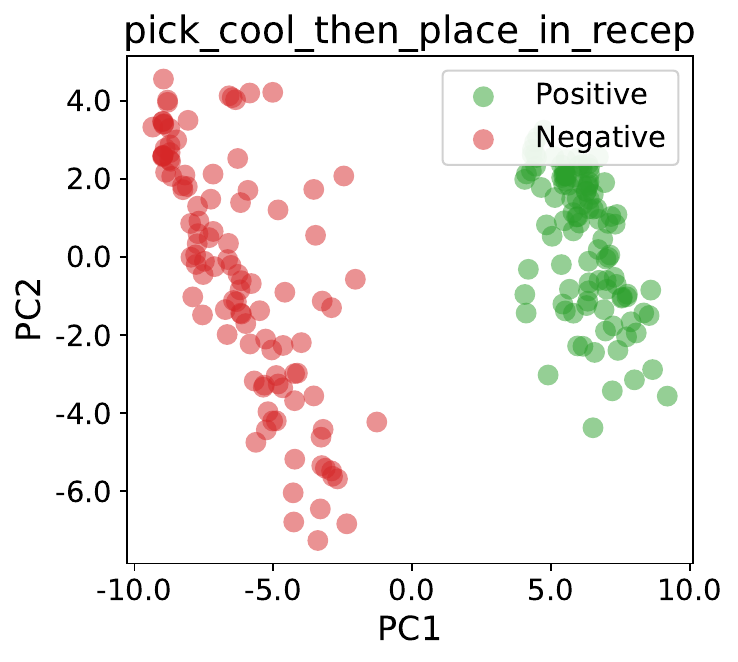}
        \caption{Intra}
        \label{fig:intra_b}
    \end{subfigure}
    \vspace{-5pt}
    \caption{PCA projections of hidden states at Layer 18 for Qwen3-4B on AlfWorld. Positive (green) and negative (red) representations exhibit distinct clustering along the primary axis of variation.}
    \vspace{-15pt}
    \label{fig:projection}
\end{figure}

Effective activation steering relies on the assumption that successful patterns and failure modes occupy distinct regions in the representation space. Projecting the hidden states from the intervention layer onto their primary axes reveals a clear geometric separation. As illustrated in Figure~\ref{fig:projection}, the projections for both inter-trajectory and intra-trajectory contrasts show that successful paths and degenerate modes form distinct clusters instead of overlapping. We also confirm this linear separability in the original high-dimensional space by training a linear classifier, which achieves high accuracy in distinguishing between the two classes (detailed in Appendix~\ref{app:svm}). This geometric separation suggests that the behavioral shift from failure to success can be projected onto a prominent direction. Extracting the first principal component of the contrastive differences helps capture this main axis of variation, mitigating some instance-specific noise.

\begin{figure}[htbp!]
    \centering
    \begin{subfigure}[b]{0.48\columnwidth}
        \centering
        \includegraphics[width=\columnwidth]{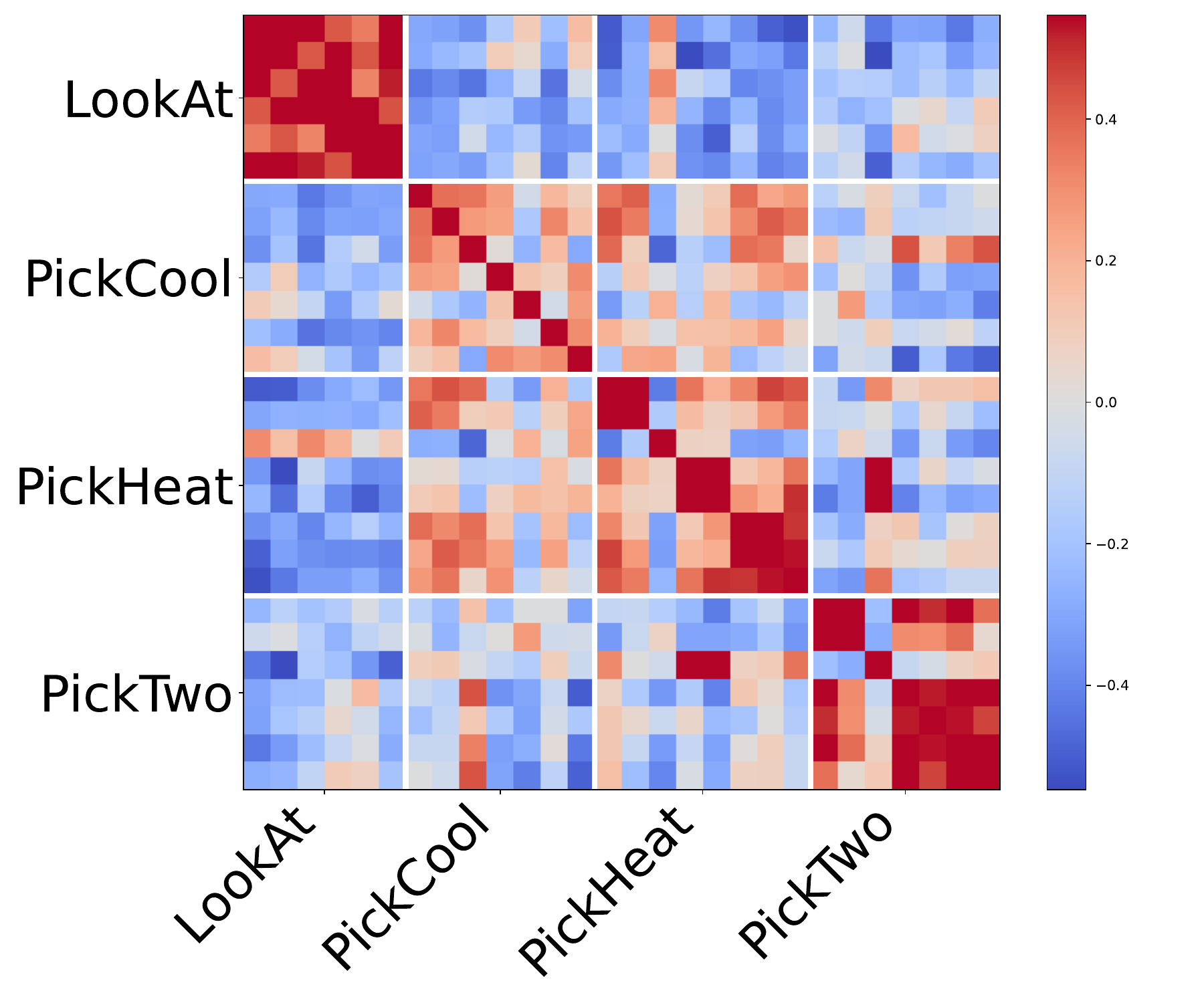}
        \caption{Inter (By Task)}
        \label{fig:sim_inter_type}
    \end{subfigure}
    \hfill
    \begin{subfigure}[b]{0.48\columnwidth}
        \centering
        \includegraphics[width=\columnwidth]{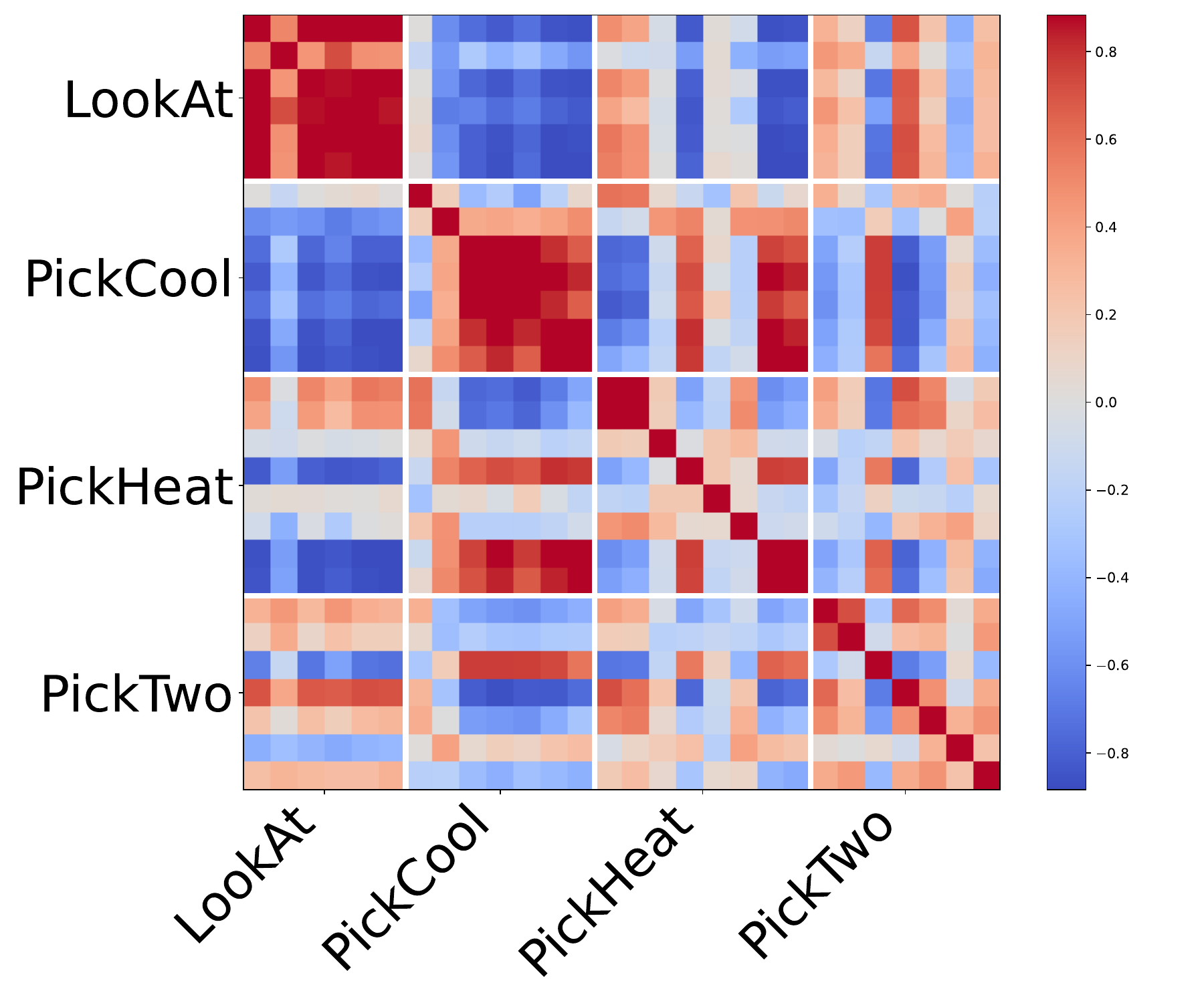}
        \caption{Intra (By Task)}
        \label{fig:sim_intra_type}
    \end{subfigure}

    \begin{subfigure}[b]{0.48\columnwidth}
        \centering
        \includegraphics[width=\columnwidth]{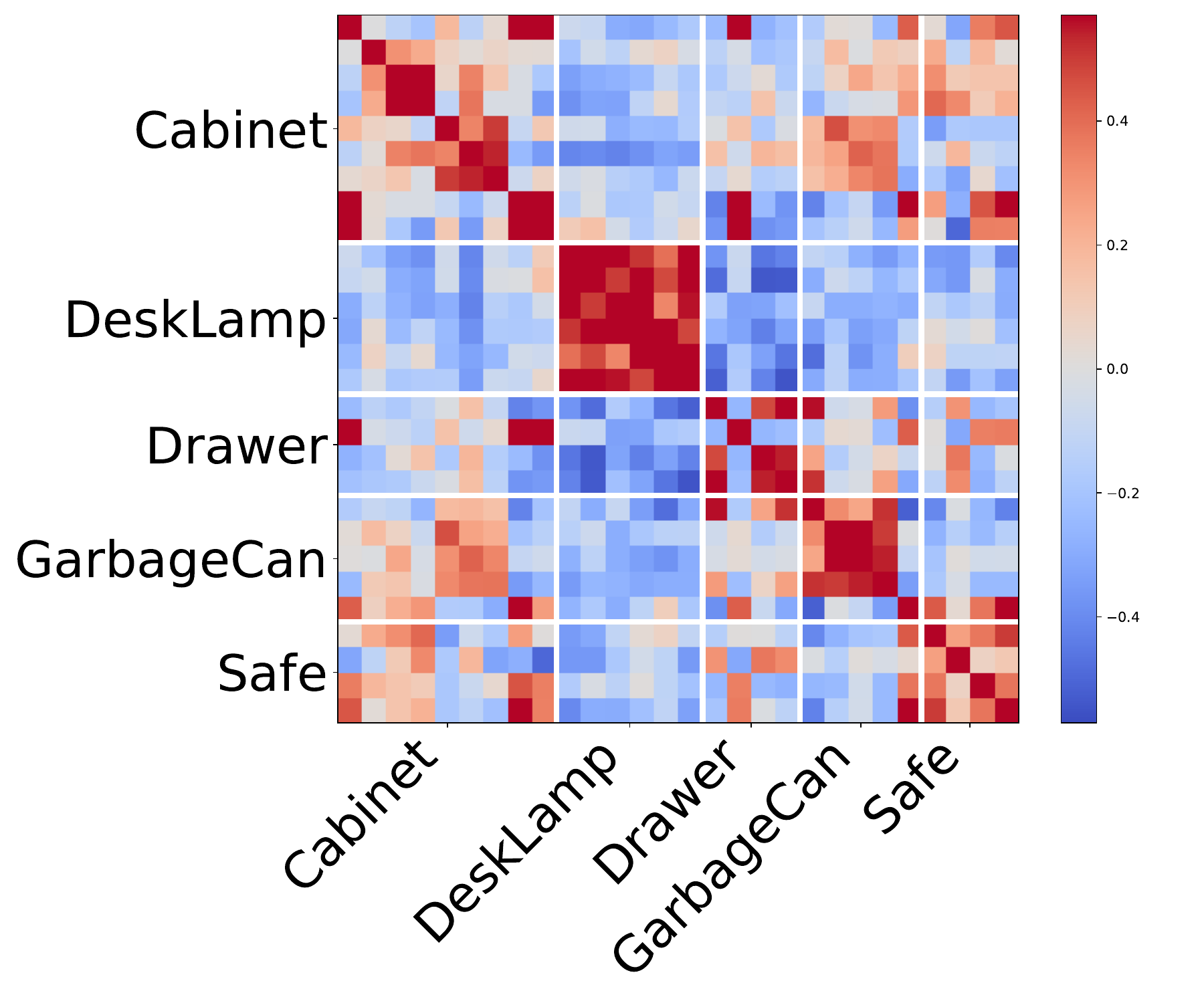}
        \caption{Inter (By Interaction)}
        \label{fig:sim_inter_target}
    \end{subfigure}
    \hfill
    \begin{subfigure}[b]{0.48\columnwidth}
        \centering
        \includegraphics[width=\columnwidth]{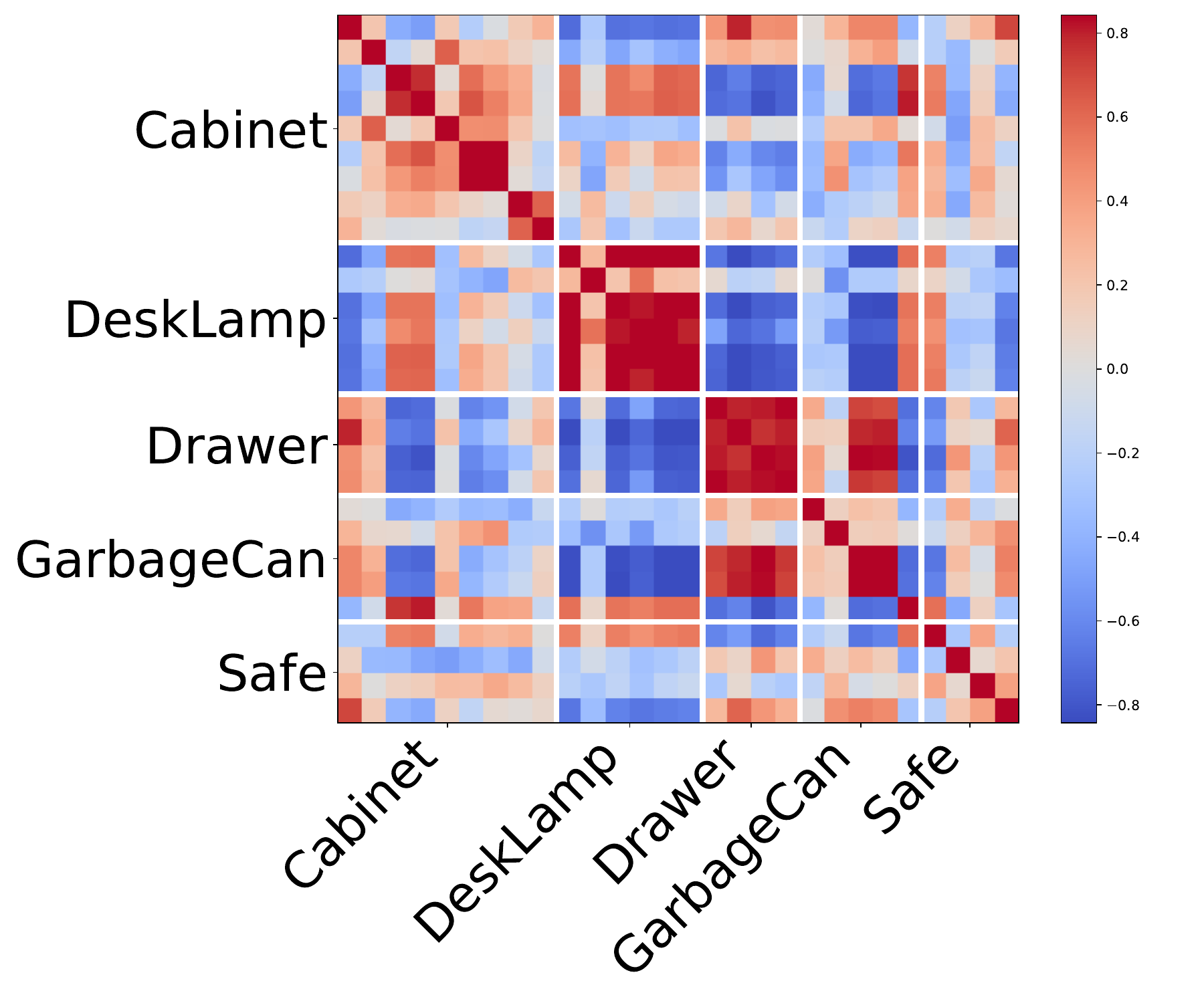}
        \caption{Intra (By Interaction)}
        \label{fig:sim_intra_target}
    \end{subfigure}
    \vspace{-8pt}
    \caption{Geometric consistency of procedural steering vectors. Heatmaps display the pairwise cosine similarity of centered vectors. (a-b) Vectors clustered by task type; (c-d) Vectors clustered by interaction target. The diagonal block structure suggests that the extracted vectors align with task-specific categories and share related behavioral features.}
    \vspace{-15pt}
    \label{fig:steering_consistency}
\end{figure}

\begin{figure*}[t]
    \centering
    \includegraphics[width=\textwidth]{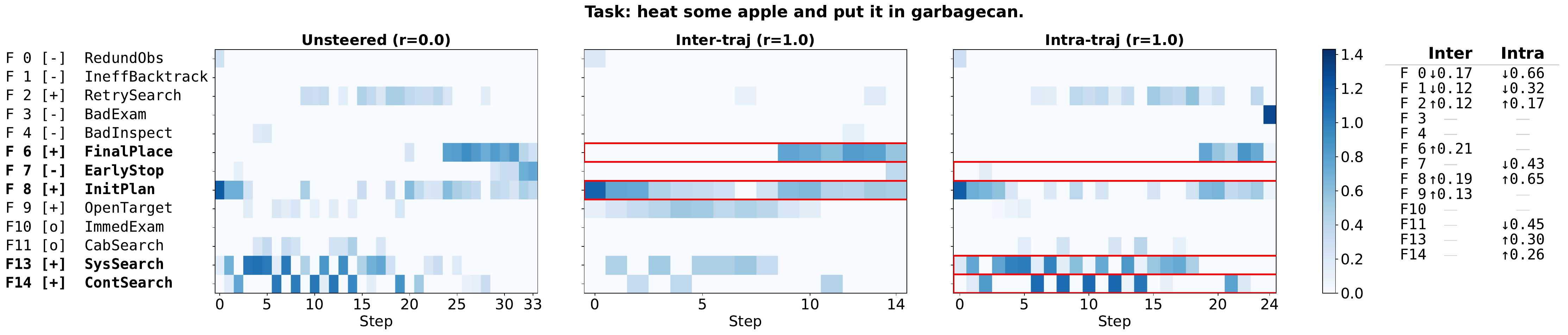}
    \vspace{-20pt}
    \caption{
        \textbf{Temporal activation of behavioral primitives across execution steps in a PickHeat task.} 
        \textbf{(Left)} The unsteered baseline exhibits an extended 33-step trajectory.
        \textbf{(Middle)} \textit{Inter-trajectory} steering is associated with a shorter trajectory by promoting early planning and object placement.
        \textbf{(Right)} \textit{Intra-trajectory} steering is observed to amplify search features and suppress premature task termination signals during execution.
    }
    \label{fig:mechanistic_steering}
    \vspace{-15pt}
\end{figure*}

\vspace{-5pt}
\subsection{Consistency of Steering Vectors}

We evaluate the structural properties of the steering vectors by computing their pairwise cosine similarities across different task instances. The heatmaps in Figures~\ref{fig:sim_inter_type} and~\ref{fig:sim_intra_type} demonstrate that these vectors maintain high internal consistency within specific task categories. Tasks such as \textit{LookAt} involve distinct interaction logic and show low similarity to the \textit{PickCool}, \textit{PickHeat}, and \textit{PickTwo} groups. These three categories display inter-task similarity alongside their high intra-task consistency because they share the fundamental \textit{Pick} sub-action. This overlapping pattern indicates the framework captures common procedural knowledge across related behaviors. The resulting diagonal block structure suggests that the memory extraction process yields representations that align with task-specific categories rather than random noise.

Grouping the vectors by specific interaction targets in Figures~\ref{fig:sim_inter_target} and~\ref{fig:sim_intra_target} highlights functional differences between the two extraction strategies. The intra-trajectory contrasts form diagonal block structures with high similarity. Local error-correction signals associated with particular items like a \textit{DeskLamp} or \textit{Drawer} remain uniform across different task instances. Correcting localized degenerate steps relies primarily on the intrinsic properties of the target object rather than the broader task context. The inter-trajectory contrasts show a less structured pattern with lower consistency. Global reasoning paths depend on the initial state of the agent, environmental layouts, and prior execution sequences. This contextual variance prevents inter-trajectory vectors from forming the distinct clusters observed in the local signals.

\begin{figure}[tbp!]
    \centering
    \includegraphics[width=0.9\columnwidth]{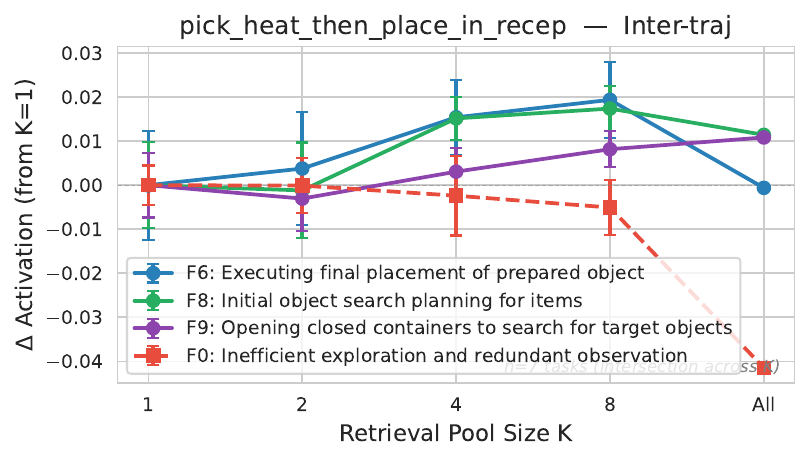}
    \vspace{-5pt}
    \includegraphics[width=0.9\columnwidth]{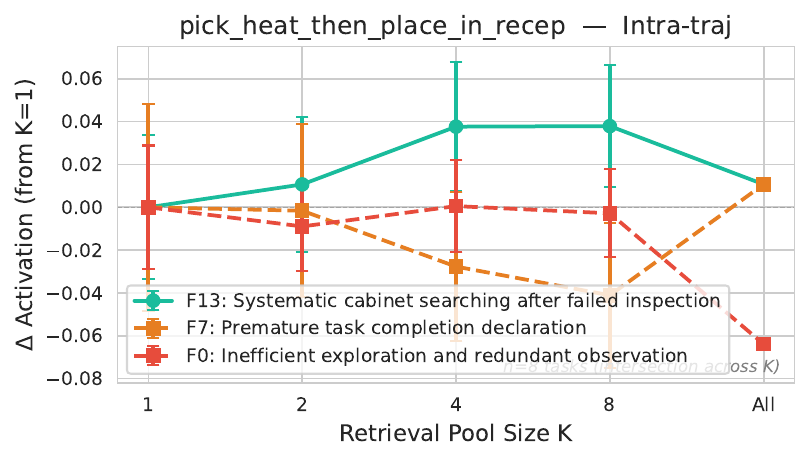}
    \vspace{-10pt}
    \caption{The impact of retrieval pool size on specific behavioral primitive activations (Solid lines for positive features; dashed lines for negative features).}
    \label{fig:retrieval_scaling_mechanics}
    \vspace{-15pt}
\end{figure}

\vspace{-5pt}
\subsection{Feature Decomposition and Interpretability of Steering Vectors}
To understand the behavioral semantics encoded within the steering vectors, we decompose the continuous representations into interpretable basis directions using sparse dictionary learning (detailed in Appendix~\ref{app:explain}). Each direction corresponds to specific behavioral primitives, which we validate by computing the mutual information between basis activations and the generated action types across the dataset. Projecting the steering vectors onto these bases reveals that the two contrastive strategies operate through distinct mechanistic pathways.

The \textit{PickHeat} task in Figure~\ref{fig:mechanistic_steering} illustrates how the two extraction strategies modulate representations through distinct mechanisms. Without intervention, the baseline model generates an extended 33-step trajectory. Applying inter-trajectory steering leads to global behavioral changes. This vector increases early planning features (F8) and sustains activations related to concrete object placement (F6), shortening the execution to 14 steps. Intra-trajectory steering targets local execution errors instead of broad execution plans. It corrects action loops by amplifying features associated with targeted searching (F13) and container interaction (F14) while suppressing signals that trigger premature task termination (F7). This difference in activation patterns suggests that the steering vectors target distinct aspects of model behavior. They systematically adjust representations to address different types of procedural errors (see more quantitative analysis in Appendix~\ref{app:feature_stat}).

\subsection{Retrieval Size and Dynamic Synthesis}

\label{sec:ablation_retrieval}
To investigate how the retrieval size affects the extracted procedural signals, we track the activation of behavioral primitives as the number of aggregated contrastive pairs increases. As shown in Figure~\ref{fig:retrieval_scaling_mechanics}, features display two distinct scaling patterns depending on their universality. For general behaviors, such as basic exploration (F9) and redundant observation (F0), activation magnitudes grow continuously as the size expands. Since these patterns are shared across most tasks, they remain prominent when aggregated across the entire dataset.
Conversely, task-specific primitives exhibit a different pattern. Features tied to explicit procedural logic, such as placing objects (F6), systematic searching (F13), or preventing premature completion (F7), reach peak activation at a moderate retrieval scale before declining. 
While minimal pools fail to separate the core procedural signal from instance-level noise, overly large pools introduce cross-task interference that suppresses specific task rules.
This observation confirms the necessity of dynamic synthesis: constraining the retrieval to a contextually relevant subset effectively balances noise reduction with signal preservation.

\vspace{-8pt}
\section{Conclusion}
\vspace{-5pt}
We propose Neural Procedural Memory (NPM), a training-free framework that represents and applies procedural behaviors through implicit activation steering. By distilling procedural knowledge from dual-granularity contrastive experiences, NPM provides an alternative paradigm for encoding memory that operates directly in the activation space. Extensive experiments show that NPM achieves performance comparable to textual baselines and provides complementary benefits when combined with explicit workflows. By utilizing the continuous activation space to encode procedural skills, NPM provides a practical foundation for managing procedural memory in LLM agents.
\section*{Limitations}
While our proposed approach demonstrates promising results, there are several aspects that could be improved. Firstly, the framework requires direct intervention in the residual stream, restricting its application to open architectures where internal activation spaces are completely accessible. Furthermore, building the contrastive repository relies on the agent occasionally generating successful trajectories, which introduces a cold-start problem in complex environments. Additionally, extracting degenerate steps relies on heuristics such as action redundancy or format invalidity. These conditions struggle to identify implicit logical fallacies without explicit environment errors. Finally, synthesized vectors compress procedural knowledge into a single representation applied constantly during generation. This static modulation lacks the flexibility to shift between behavioral primitives across different execution stages. Exploring dynamic interventions that adapt throughout task execution is a direction for future work.

% \section*{Acknowledgments}

% This document has been adapted
% by Steven Bethard, Ryan Cotterell and Rui Yan
% from the instructions for earlier ACL and NAACL proceedings, including those for
% ACL 2019 by Douwe Kiela and Ivan Vuli\'{c},
% NAACL 2019 by Stephanie Lukin and Alla Roskovskaya,
% ACL 2018 by Shay Cohen, Kevin Gimpel, and Wei Lu,
% NAACL 2018 by Margaret Mitchell and Stephanie Lukin,
% Bib\TeX{} suggestions for (NA)ACL 2017/2018 from Jason Eisner,
% ACL 2017 by Dan Gildea and Min-Yen Kan,
% NAACL 2017 by Margaret Mitchell,
% ACL 2012 by Maggie Li and Michael White,
% ACL 2010 by Jing-Shin Chang and Philipp Koehn,
% ACL 2008 by Johanna D. Moore, Simone Teufel, James Allan, and Sadaoki Furui,
% ACL 2005 by Hwee Tou Ng and Kemal Oflazer,
% ACL 2002 by Eugene Charniak and Dekang Lin,
% and earlier ACL and EACL formats written by several people, including
% John Chen, Henry S. Thompson and Donald Walker.
% Additional elements were taken from the formatting instructions of the \emph{International Joint Conference on Artificial Intelligence} and the \emph{Conference on Computer Vision and Pattern Recognition}.

\bibliography{main}

\appendix
\section{Implementation Details}
\label{app:implementation}

\subsection{Dataset Details}

Our evaluation utilizes four simulated environments that test diverse procedural skills and long-horizon planning capabilities. 

ALFWorld~\cite{ALFWorld} is a text-based embodied environment where agents navigate simulated rooms and interact with household objects to complete daily tasks. WebShop~\cite{WebShop} provides a simulated e-commerce website where agents must execute search queries, navigate product pages, and select specific item attributes to purchase products matching detailed user instructions. ScienceWorld~\cite{wang2022scienceworld} simulates an elementary school science laboratory, requiring agents to follow strict procedural logic to conduct experiments. BabyAI~\cite{chevalier-boisvert2018babyai} is a partially observable gridworld environment where agents execute sequential movement and manipulation commands to complete specified spatial tasks.

\subsection{Data Construction and Model Configuration}

We construct the contrastive dataset by sampling tasks directly from the training splits of ALFWorld, WebShop, ScienceWorld, and BabyAI. For each training task, we generate multiple trajectories using the target model configured with specific sampling parameters. 

To isolate degenerate behaviors for intra-trajectory contrasts, we apply deterministic rules based on action redundancy and invalidity. Redundancy is identified through pattern matching that detects repetitive loops or consecutive identical action sequences. Categorizing segments where the agent enters a state unable to progress normally as degenerate steps provides a reasonable heuristic for capturing procedural failures. Invalidity is determined by matching environment-specific error messages and format violations, which filters irrational actions from the effective trajectory pool.

All experiments employ \texttt{sentence-transformers/all-mpnet-base-v2} for dense retrieval to retrieve a fixed number of the most relevant historical tasks. The agent then performs greedy decoding to ensure the reproducibility of the evaluation results. The detailed sampling parameters and retrieval configurations for all evaluated environments are summarized in Table \ref{tab:implementation_details}.

\begin{table}[h]
\centering
\caption{Data construction parameters. All parameters except the maximum interaction steps are kept consistent across the four evaluated benchmarks.}
\begin{tabular}{lc}
\toprule
Parameter & Value \\
\midrule
Sampled trajectories ($N$) & 16 \\
Temperature & 0.7 \\
Top-$p$ & 0.8 \\
Top-$k$ & 20 \\
Max turns (ALFWorld) & 50 \\
Max turns (WebShop) & 15 \\
Max turns (ScienceWorld) & 30 \\
Max turns (BabyAI) & 20 \\
Retrieval count ($K$) & 8 \\
\bottomrule
\end{tabular}
\label{tab:implementation_details}
\end{table}

\subsection{Dynamic Synthesis and Vector Extraction}
\label{app:synthesis_strategy}

To determine the intervention granularity $g$, we estimate task complexity using the average execution length of the retrieved successful trajectories, denoted as $L_q$. We compare this length against the global average trajectory length of the respective benchmark, denoted as $\bar{L}$. Tasks with extended horizons typically encounter localized procedural failures such as repetitive loops, making the concentrated error-correction of intra-trajectory steering more appropriate. Conversely, shorter tasks typically fail due to misaligned global planning, which benefits from the broader behavioral alignment provided by inter-trajectory steering. The system selects the appropriate granularity based on a simple thresholding strategy, with the scaling factor $\gamma$ set to 1.5 in our experiments:

\begin{equation}
g = 
\begin{cases} 
\text{intra-trajectory}, & \text{if } L_q > \gamma \bar{L} \\
\text{inter-trajectory}, & \text{otherwise}
\end{cases}
\end{equation}

Nevertheless, identifying the optimal granularity for every instance remains a non-trivial challenge. 
Our heuristic selection occasionally scores lower than the oracle single-granularity baseline in specific settings. Future research could explore more effective methods for allocating steering vectors.

After selecting the appropriate contrastive pool, we extract the steering vectors using a standard principal component analysis pipeline. The paired positive and negative activations are first mean-centered to eliminate shared task-specific biases. We then compute the principal components of these normalized representations and extract the dominant component as the final continuous steering vector.

\subsection{Steering Configuration}
We apply activation steering to the middle-to-late layers of the language models. Specifically, we inject the steering vectors into layers 17 to 19 for Qwen3-4B and Qwen3-8B, and layers 46 to 48 for MiniCPM3-4B. 

To achieve dynamic steering across these targeted layers, we adapt a simplified version of the calibration mechanism based on Kullback-Leibler divergence proposed by~\citet{scalena-etal-2024-multi}. This method tracks the probability shift caused by the steering vector and reduces the intervention strength if the modified next-token distribution $P_\alpha(\cdot \mid x_{<t})$ deviates from the original unsteered distribution $P(\cdot \mid x_{<t})$ by a large margin. We define a discrete set of candidate scales $\mathcal{A}$ and select the optimal strength $\alpha^*$ by identifying the maximum value that keeps the divergence below a target threshold $\epsilon$:

\begin{equation}
\begin{aligned}
\alpha^* = \max \Big\{ \alpha \in \mathcal{A} \;\Big|\; & D_{\text{KL}} \big( P(\cdot \mid x_{<t}) \\
&\parallel P_\alpha(\cdot \mid x_{<t}) \big) \le \epsilon \Big\}
\end{aligned}
\end{equation}

\section{Steering Vector Interpretability and Feature Analysis Details}
\label{app:explain}

\subsection{Feature Extraction and Semantic Annotation}

To investigate the underlying mechanisms of implicit activation steering, we decompose the continuous representation space into interpretable basis directions following the methodology of~\citet{venhoff2025basemodelsknowreason}. 
We collect step-level hidden states from historical interaction trajectories at layer 18 for the Qwen3-4B model and apply a sparse dictionary learning algorithm with a top-k activation constraint. The dictionary size is constrained to 16 dimensions with a sparsity threshold of 3. This dimensional configuration aligns with prior research on isolating specific high-level procedural behaviors. The empirical observation of inactive dimensions within this learned dictionary indicates that a 16-dimensional space provides adequate capacity for capturing the procedural variations specific to agentic task-solving steps.

We assign interpretable semantics to these learned basis directions by analyzing their activation patterns across the dataset. We quantify behavioral selectivity by calculating the mutual information between the activation state of each basis direction $f$ and the discrete action type $a$ generated by the agent. This statistical measurement is defined as:
\begin{equation}
\begin{aligned}
MI(F; A) = \sum_{a} \Biggl[ & P(f,a) \log \frac{P(f,a)}{P(f)P(a)} \\
& + P(\neg f,a) \log \frac{P(\neg f,a)}{P(\neg f)P(a)} \Biggr]
\end{aligned}
\end{equation}

Building on established automated interpretability techniques~\cite{bills2023language,paulo2025automatically}, we utilize DeepSeek-V3.2 to analyze these statistical distributions alongside top-activating examples. The model processes this information according to the prompt template detailed in Figure~\ref{fig:prompt_template} to generate descriptive textual labels and determine the behavioral polarity for each active basis direction. The resulting semantic annotations for the ALFWorld benchmark are summarized in Table~\ref{tab:feature_labels}.

\begin{table*}[t]
\centering
\caption{Automated annotations and activation statistics of the learned basis directions in the ALFWorld environment.}
\begin{tabularx}{\textwidth}{c X cccc}
\toprule
Index & Semantic Label & 
\begin{tabular}{@{}c@{}}Overall \\ Active\end{tabular} & 
\begin{tabular}{@{}c@{}}Active on \\ Success\end{tabular} & 
\begin{tabular}{@{}c@{}}Active on \\ Failure\end{tabular} & 
Polarity \\
\midrule
0 & \textbf{RedundObs}: Inefficient exploration and redundant observation & 42.0\% & 7.3\% & 46.2\% & Negative \\
1 & \textbf{IneffBacktrack}: Ineffective backtracking after failed object placement & 10.0\% & 0.5\% & 11.1\% & Negative \\
2 & \textbf{RetrySearch}: Searching for objects after failed container checks & 17.9\% & 24.3\% & 17.1\% & Positive \\
3 & \textbf{BadExam}: Misguided final-step execution with examination tasks & 1.5\% & 0.8\% & 1.6\% & Negative \\
4 & \textbf{BadInspect}: Misguided visual inspection of desk objects & 23.5\% & 7.7\% & 25.4\% & Negative \\
5 & \textbf{Dead}: Dead feature & 0.0\% & 0.0\% & 0.0\% & N/A \\
6 & \textbf{FinalPlace}: Executing final placement of prepared object & 23.1\% & 33.0\% & 21.9\% & Positive \\
7 & \textbf{EarlyStop}: Premature task completion declaration & 36.5\% & 8.0\% & 40.0\% & Negative \\
8 & \textbf{InitPlan}: Initial object search planning for food items & 32.6\% & 67.1\% & 28.4\% & Positive \\
9 & \textbf{OpenTarget}: Opening closed containers to search for target objects & 20.9\% & 22.7\% & 20.7\% & Positive \\
10 & \textbf{ImmedExam}: Immediate visual examination of target objects & 6.4\% & 6.3\% & 6.5\% & Neutral \\
11 & \textbf{CabSearch}: Opening closed cabinets to search for objects & 22.1\% & 24.7\% & 21.8\% & Neutral \\
12 & \textbf{Dead}: Dead feature & 0.0\% & 0.0\% & 0.0\% & N/A \\
13 & \textbf{SysSearch}: Systematic cabinet searching after empty inspection & 41.2\% & 60.0\% & 38.9\% & Positive \\
14 & \textbf{ContSearch}: Opening closed containers to search for objects & 22.3\% & 37.6\% & 20.4\% & Positive \\
15 & \textbf{Dead}: Dead feature & 0.0\% & 0.0\% & 0.0\% & N/A \\
\bottomrule
\end{tabularx}
\label{tab:feature_labels}
\end{table*}

\begin{figure*}[t]
\begin{tcolorbox}[title=\textbf{Prompt Template for Automated Feature Annotation}, colback=gray!5!white, colframe=gray!75!black, fonttitle=\bfseries, fontupper=\small]
We are analyzing features learned by a sparse dictionary trained on the hidden states of an LLM agent solving interactive tasks.

\vspace{0.5em}
\textbf{Environment:} ALFWorld is a text-based household environment where an agent must complete tasks like finding objects, cleaning them, heating/cooling them, and placing them in target locations. The agent navigates rooms, opens containers, picks up objects, and uses appliances through text commands.

\vspace{0.5em}
Each "feature" is a direction in the model's representation space that activates on specific behavioral patterns. Your job is to identify what behavioral pattern causes Feature [ID] to activate, and whether it is associated with effective or ineffective agent behavior.

\vspace{0.5em}
\#\# Top Activating Examples (sorted by activation strength)

\#\#\# Example 1 (activation=[val], reward=[val])

Observation: [text]

Action: [text]

[... additional examples ...]

\#\# Non-Activating Examples (Feature [ID] does NOT activate on these)

\#\#\# Example 1 (reward=[val])

Observation: [text]

Action: [text]

[... additional examples ...]

\vspace{0.5em}
\#\# Outcome Statistics

- This feature activates in [X]\% of all steps

- Activation rate on successful trajectories: [X]\%

- Activation rate on failed trajectories: [X]\%

- Mean activation strength on successful trajectories: [val]

- Mean activation strength on failed trajectories: [val]

\vspace{0.5em}
\#\# Action Type Distribution (when this feature is active)

The following shows ALL action types in this environment, sorted by how much more likely they are to co-occur with this feature compared to the overall action distribution (ratio > 1 means over-represented, < 1 means under-represented):

- [action type]: [ratio]x (P(active|action)=[X]\%, mean\_activation=[val])

[... additional action types ...]

\vspace{0.5em}
\#\# Instructions
Based on ALL the evidence above (examples, outcome statistics, and action distribution), provide a comprehensive interpretation of this feature:

1. Identify the common behavioral pattern across the activating examples that distinguishes them from the non-activating examples.

2. Provide a short label (3-8 words in English) for this feature. The label should reflect both WHAT the behavior is and WHETHER it contributes to task success or failure.

3. Provide a one-paragraph explanation of what this feature detects, integrating evidence from the examples, outcome statistics, and action type distribution.

4. Classify the feature's polarity: "positive" if it is associated with effective/successful behavior, "negative" if associated with ineffective/failing behavior, or "neutral" if no clear association.

5. Rate your confidence (high/medium/low) in this interpretation.

\vspace{0.5em}
Respond in JSON format:

\{"label": "...", "explanation": "...", "polarity": "positive | negative | neutral", "confidence": "high | medium | low", "key\_evidence": ["...", "..."]\}
\end{tcolorbox}
\caption{Prompt template for automated interpretation of geometric basis directions.}
\label{fig:prompt_template}
\end{figure*}

\subsection{Geometric Projection of Steering Vectors}

We analyze the functional composition of the extracted principal component steering vectors by projecting them onto the annotated basis directions. The geometric projection is computed as the dot product between the normalized steering vector $\mathbf{v}$ and each unit-norm basis vector $\mathbf{w}_i$ derived from the dictionary learning phase:

\begin{equation}
\label{equ:9}
p_i = \mathbf{w}_i^\top \frac{\mathbf{v}}{\|\mathbf{v}\|}
\end{equation}

A positive projection value $p_i > 0$ indicates that the steering intervention amplifies the corresponding behavioral primitive. A negative value $p_i < 0$ signifies suppression of that specific behavior. This mathematical decomposition confirms that the steering vectors function as composite modifiers that simultaneously inhibit pathological patterns and stimulate necessary subsequent steps to correct agent reasoning.

\section{Quantitative Validation of Methodology}
\subsection{Linear Separability in High-Dimensional Manifold}
\label{app:svm}
We train a linear support vector machine on the hidden states extracted from the target intervention layers of the retrieved trajectories to verify the linear separability of successful and failed procedural logic. This evaluation determines whether effective and degenerate reasoning modes occupy distinct regions within the representation space~\cite{marks2024massmean, zou2023transparency}. We perform 5-fold stratified cross-validation on the Qwen3-4B model across all four evaluated benchmarks. The classification accuracies at both the inter-trajectory and intra-trajectory levels are reported in Table~\ref{tab:svm_separability}. The consistently high classification accuracy across different environments confirms that a linear hyperplane effectively divides the two reasoning classes. This mathematical separability provides the theoretical justification for applying principal component analysis to extract a dominant directional vector for behavioral modulation.

\begin{table}[h]
\centering
\caption{Linear SVM classification accuracy (\%) on the native high-dimensional hidden states of Qwen3-4B across 5-fold stratified cross-validation.}
\label{tab:svm_separability}
\begin{tabular}{lcc}
\toprule
Benchmark & Inter-traj. & Intra-traj. \\
\midrule
ALFWorld     & 99.55 ($\pm$0.51) & 99.99 ($\pm$0.01) \\
WebShop      & 88.46 ($\pm$0.55) & 99.91 ($\pm$0.10) \\
ScienceWorld & 93.41 ($\pm$0.69) & 99.81 ($\pm$0.12) \\
BabyAI       & 99.53 ($\pm$0.23) & 99.61 ($\pm$0.10) \\
\bottomrule
\end{tabular}
\end{table}
\subsection{Accuracy of Heuristic Rules for Degenerate Steps}
\label{app:heuristic_rules}
The efficacy of intra-trajectory steering depends on isolating degenerate steps without contaminating the negative representation pool. We evaluate the reliability of our deterministic isolation rules based on consecutive loop detection and invalid format matching by analyzing sampled steps from 1,000 historical trajectories per benchmark. An LLM-based evaluator assesses each step against the criteria of action redundancy and environment constraints to compute the false positive and false negative rates. The evaluation metrics presented in Table~\ref{tab:heuristic_accuracy} show a low overall false positive rate of 0.61\% and a low overall false negative rate of 6.34\%, indicating that the heuristics reliably isolate genuine procedural failures and maintain high purity in the negative representations. The minor variance in error rates across benchmarks, such as the slightly higher false negative rate of 9.30\% in WebShop, occurs because rigid deterministic rules can occasionally miss subtle, domain-specific logical errors. These undetected failures subsequently mix into the effective step pool. Although this introduces noise into the positive representations, the empirical results confirm that the overall contrastive signal remains robust for behavioral modulation. Future work could address this limitation by replacing heuristic rules with more capable evaluation mechanisms, such as utilizing a language model as a judge to identify nuanced reasoning failures during data construction.

\begin{table}[ht]
\centering
\small
\caption{Evaluation of heuristic rule accuracy based on step-level annotation across four benchmarks. False positive rate (FPR) and false negative rate (FNR) are evaluated by an LLM-as-a-judge on sampled steps from 1,000 trajectories per environment.}
\label{tab:heuristic_accuracy}
\resizebox{\columnwidth}{!}{%
\begin{tabular}{lcc}
\toprule
\textbf{Benchmark} & \textbf{FPR} & \textbf{FNR} \\
\midrule
ALFWorld & 0.00\% (0/2,862) & 3.57\% (106/2,967) \\
WebShop & 0.64\% (16/2,490) & 9.30\% (268/2,883) \\
ScienceWorld & 0.04\% (1/2,833) & 6.97\% (184/2,640) \\
BabyAI & 1.75\% (51/2,908) & 5.73\% (171/2,983) \\
\midrule
Overall & 0.61\% (68/11,093) & 6.34\% (729/11,473) \\
\bottomrule
\end{tabular}%
}
\end{table}

\subsection{Distribution Analysis of Inter- and Intra-Trajectory Feature Activations}
\label{app:feature_stat}
To evaluate how steering vectors modulate behaviors, we analyze their geometric distribution across the basis directions defined in Equation~\ref{equ:9} using normalized entropy and top-three concentration. For any task, the projection proportion along each basis direction is:
\begin{equation}
\tilde{p}_i = \frac{|p_i|}{\sum_{j=1}^N |p_j|}
\end{equation}
The global uniformity of this distribution is captured by the normalized entropy:
\begin{equation}
H_{\text{norm}} = -\frac{\sum_{i=1}^N \tilde{p}_i \log \tilde{p}_i}{\log N}
\end{equation}
where $N$ denotes the total number of active directions. To measure local focus, we also compute the top-three concentration, representing the cumulative proportion of the three largest absolute projection values:
\begin{equation}
C_{\text{top3}} = \frac{\sum_{i \in \Omega} |p_i|}{\sum_{j=1}^N |p_j|}
\end{equation}
with $\Omega$ denoting the indices of these three dominant directions.

\begin{figure}[htbp]
    \centering
    \begin{minipage}{0.9\columnwidth}
        \centering
        \includegraphics[width=\columnwidth]{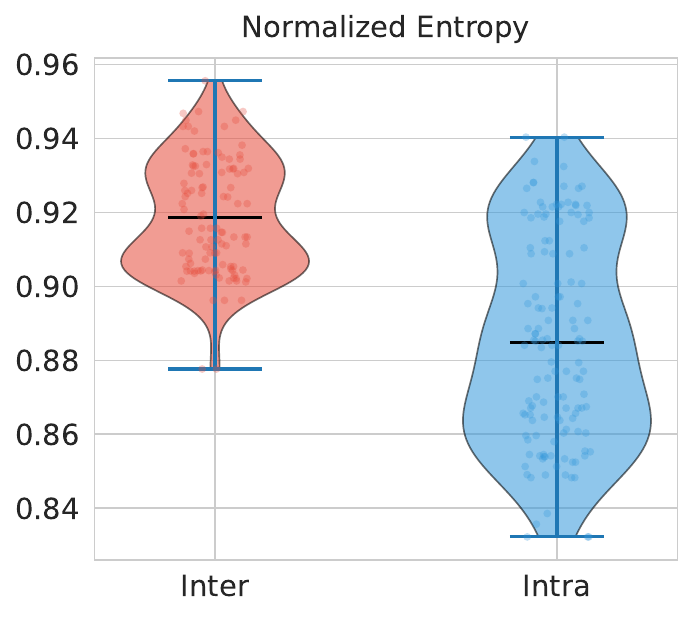} 
        \caption{Normalized entropy for inter-trajectory and intra-trajectory steering vectors across the ALFWorld benchmark.}
        \label{fig:normalized_entropy}
    \end{minipage}
    \hfill
    \begin{minipage}{0.9\columnwidth}
        \centering
         \includegraphics[width=\columnwidth]{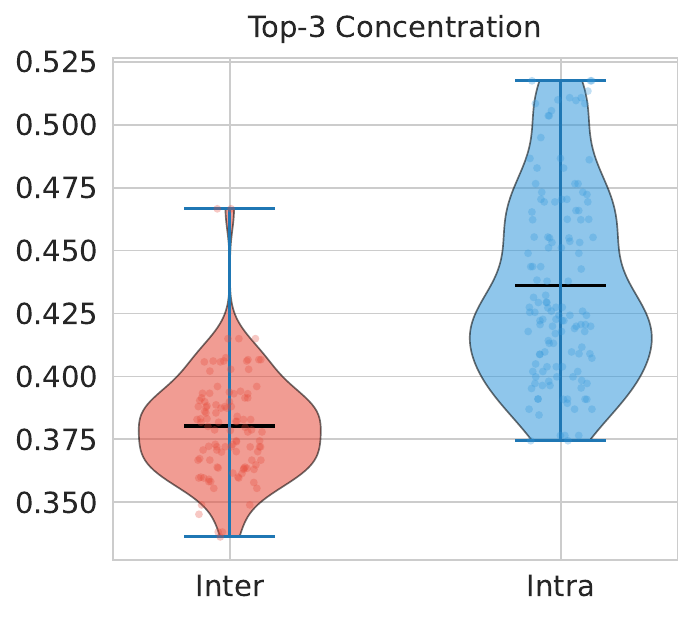}
        \caption{Top-3 concentration for inter-trajectory and intra-trajectory steering vectors across the ALFWorld benchmark.}
        \label{fig:top3_concentration}
    \end{minipage}
\end{figure}

Empirical analysis of these metrics reveals distinct structural differences between the two steering granularities. As shown in Figure~\ref{fig:normalized_entropy}, inter-trajectory steering vectors exhibit a mean normalized entropy of 0.919, whereas intra-trajectory vectors yield a lower mean of 0.885. This difference indicates that macro-level planning steering operates via a more distributed modulation across multiple basis directions. Conversely, Figure~\ref{fig:top3_concentration} shows that the mean top-three concentration is 38.0\% for inter-trajectory steering and 43.6\% for intra-trajectory steering. The higher concentration of intra-trajectory vectors indicates a more targeted activation pattern. Instead of adjusting the overall reasoning flow, intra-trajectory intervention selectively modulates specific behavioral dimensions to resolve localized failures. These patterns suggest that global planning guidance tends to rely on distributed representations, whereas local error recovery is associated with targeted corrections.

\section{System Efficiency and Overhead}
\label{app:system_efficiency}
\subsection{Storage and Computation Overhead}
\label{app:storage_compute}
The framework introduces minimal storage overhead compared to text-based retrieval. For each historical trajectory, the system extracts three representations at each targeted layer: one last-token state for inter-trajectory contrast and two mean-pooled states representing effective and degenerate steps for intra-trajectory contrast. Given a hidden dimension $d$, floating-point byte size $b$, and $L$ target layers, the total storage requirement per trajectory is $3 \times L \times d \times b$. For Qwen3-4B with $d=2560$ in half-precision format ($b=2$) targeting three layers, the footprint is 45 kilobytes. This fixed constraint ensures that storage scales independently of the original interaction length. Unlike traditional retrieval systems that consume substantial video memory by expanding the key-value cache with extensive textual contexts, activation steering modifies the residual stream directly to bypass this limitation. The offline computation for memory construction is limited to a single forward pass to extract representations. These continuous vectors can be efficiently managed in a standard database without requiring model parameter updates.

\subsection{Latency Analysis}
\label{app:latency}

We evaluate the inference latency of the proposed framework across retrieval, scale selection probing, vector synthesis, prompt prefill, and autoregressive decoding. The timing results in Table~\ref{tab:latency} indicate that textual memory incurs a substantial prefill overhead, increasing latency from 63.46 ms to 279.89 ms, as the self-attention mechanism must process extended textual contexts. In contrast, NPM maintains a prefill latency of 71.09 ms, nearly identical to the No Memory baseline. Although NPM introduces additional phases to compute steering vectors and steering strength, this process imposes no extra latency penalty relative to textual memory. During autoregressive decoding, vector injection involves only an element-wise addition within the residual stream. This basic operation adds no measurable delay to token generation. These efficiency characteristics establish implicit activation steering as a practical and scalable paradigm for persistent agentic memory.
\begin{table}[htbp]
\centering
\small
\caption{Inference latency breakdown (ms).}
\label{tab:latency}
\resizebox{\columnwidth}{!}{%
\begin{tabular}{lcccc}
\toprule
Method & Retrieval & Synthesis & Probe & Prefill \\
\midrule
No Memory      & ---   & ---   & ---   & 63.46 \\
Textual Memory & 13.83 & ---   & ---   & 279.89 \\
NPM (Ours)     & 13.83 & 77.42 & 130.46 & 71.09 \\
\bottomrule
\end{tabular}%
}
\end{table}
\section{Examples and Case Studies}
\subsection{Examples of Textual Procedural Memory}
\label{app:example}
Examples are provided in Figure~\ref{fig:webshop_insights},~\ref{fig:alfworld_workflows}.
% ==========================================
% WebShop Example
% ==========================================
\begin{figure*}[t]
\begin{tcolorbox}[title=\textbf{Example of Insights Baseline}, colback=gray!5!white, colframe=gray!75!black, fonttitle=\bfseries]
\textbf{Task Instruction:} \\
\textit{``i am looking for wild caught, ready to eat sardines in a tomato sauce, and price lower than 50.00 dollars''}

\textbf{Insights:}
\begin{itemize}
    \item Ensure the product's title and description explicitly mention the requested item type, size, color, and price range to avoid misselection.
    \item Always verify that the product's price is clearly stated and falls within the user's specified budget range to avoid overpayment.
    \item Prioritize products with clear and detailed descriptions that explicitly confirm the product meets all requested specifications, including intended use, size, and color.
    \item Ensure the product's size is explicitly mentioned in both the title and description and matches the exact requested dimensions to avoid purchasing an incorrect product.
    \item Always confirm that the product's title and description explicitly mention the requested use case, color, and price to ensure alignment with the user's requirements.
    \item Verify that the product's key features explicitly include the requested functionality (e.g., wireless, battery inclusion) in both the title and description.
    \item Prioritize products where the requested attributes (such as type, size, and price) are explicitly mentioned in both the title and description to minimize ambiguity.
    \item Always check the product's price in both the title and description before proceeding to purchase to avoid unexpected costs.
    \item[\dots] \textit{(total 20 insights provided in context.)}
\end{itemize}
\end{tcolorbox}
\caption{Example of Insights Baseline for webshop}
\label{fig:webshop_insights}
\end{figure*}
% ==========================================
% ALFWorld Example
% ==========================================
\begin{figure*}[t]
\begin{tcolorbox}[title=\textbf{Example of Workflows Baseline}, colback=gray!5!white, colframe=gray!75!black, fonttitle=\bfseries]
\textbf{Task Goal:} \\
\textit{put a clean soapbar in countertop.}

\textbf{Workflows:}
\begin{itemize}
    \item find\_and\_take\_object \\    
When you need to find and take an object, search in likely locations.
    \begin{itemize}
        \item go to \{likely\_location\}
        \item{} [if container is closed] open \{container\}
        \item take \{object\} from \{location\}
    \end{itemize}
  \item open\_and\_search\_container \\  
When you need to open a container and search inside.  
\begin{itemize}  
    \item go to \{container\}  
    \item open \{container\}  
    \item{} [check contents]  
\end{itemize}  

\item search\_multiple\_locations \\  
When you need to search multiple locations for an object.  
\begin{itemize}  
    \item go to \{location 1\}  
    \item go to \{location 2\}  
    \item go to \{location 3\}  
    \item{} [repeat as needed]  
\end{itemize}  

\item clean\_and\_place\_object \\  
When you need to clean an object and place it in a specific location.  
\begin{itemize}  
    \item go to \{cleaning\_location\}  
    \item clean \{object\} with \{cleaning\_location\}  
    \item go to \{target\_location\}  
    \item move \{object\} to \{target\_receptacle\}  
\end{itemize}
\item[\dots]
\end{itemize}
\end{tcolorbox}
\caption{Example of Workflows Baseline for AlfWorld}
\label{fig:alfworld_workflows}
\end{figure*}
\subsection{Case Study: Alleviating Text-Action Disconnect in Action Execution}
\label{app:case_study}
We examine agent behavior in a multi-object retrieval task requiring the placement of two CDs into a safe (Table~\ref{tab:workflow_vs_npm}). The baseline model receives an explicit textual workflow outlining the necessary sequential operations and repetition conditions. When the environment rejects an attempt to pick up a second object due to inventory constraints, the text-augmented agent fails to update its internal state representation. The model generates subsequent commands based on an incorrect assumption of successful object acquisition and enters an invalid action loop of attempting to place an unpossessed item into the receptacle. This behavior demonstrates the text-action disconnect where semantic comprehension of discrete textual rules does not ensure sustained state maintenance during execution. The agent equipped with neural procedural memory completes the entire fetch-and-place cycle without textual prompting. Intervening directly in the residual stream allows the synthesized steering vectors to suppress neural activations associated with redundant loops while amplifying features related to sequential planning as identified in prior mechanistic analyses. Modulating the continuous activation space directly enforces the procedural logic required for multi-step execution and prevents the trajectory from degrading into the repetitive errors observed when relying solely on external textual memory.

\begin{table*}
\centering
\caption{Qualitative comparison of procedural memory in the ``find two cd and put them in safe" task. Despite the \textbf{Textual Workflow} providing explicit instructions, the agent fails to manage the multi-step trip after an initial action failure. \textbf{NPM} steers the agent implicitly adhering to the sequential requirements.}
\label{tab:workflow_vs_npm}
\small  
\begin{tabularx}{\textwidth}{l|X|X}
\toprule
\textbf{Task} & \multicolumn{2}{c}{\textbf{Your task is to: find two cd and put them in safe.}} \\ \midrule
\textbf{Memory} & \textbf{Textual Workflow:} \newline \textit{``[Workflow: search\_and\_place\_multiple\_objects] ... \textbf{take \{object 1\}} from \{location\} ... \textbf{go to \{target\_receptacle\}} ...''} & \textbf{NPM (Ours):} \newline \textit{Implicit steering vector distilled from historical multi-object retrieval tasks.} \\ \midrule
\textbf{Discovery} & \multicolumn{2}{l}{Both agents successfully locate two target CDs in the environment and attempt to pick them up.} \\ \midrule
\textbf{Constraint Hit} & \textbf{Action:} \texttt{take cd 1} $\rightarrow$ \texttt{go to desk 2} $\rightarrow$ \texttt{take cd 2} \newline \textbf{Obs:} (Action rejected as hands are full). \newline \textbf{Thought:} ``I have taken cd 2 from desk 2. Now I need to go to the safe to place both cds.'' \textit{\color{red}(Erroneous state tracking)} & \textbf{Action:} \texttt{take cd 3} $\rightarrow$ \texttt{go to desk 1} $\rightarrow$ \texttt{take cd 1} \newline \textbf{Obs:} (Action rejected as hands are full). \newline \textbf{Thought:} ``I have found the second cd (cd 1) on desk 1. Now I need to go to the safe and put both cds into it.'' \textit{\color{blue}(Correct multi-step planning)} \\ \midrule
\rowcolor[HTML]{FFF2CC} 
\textbf{Procedural Divergence} & \textbf{Thought:} ``I need to move cd 2 into the safe... I will move cd 2 into it.'' \newline \textbf{Action:} \texttt{move cd 2 to safe 1} \newline \textit{\color{red}(Hallucination)} & \textbf{Thought:} ``I have placed the first cd (cd 3) into the safe... Now I need to find the second cd (cd 1)... \textbf{I will go to desk 1 to retrieve the second cd}.'' \newline \textbf{Action:} \texttt{go to desk 1} \\ \midrule
\textbf{Execution} & \textbf{Action:} \texttt{move cd 2 to safe 1} $\rightarrow$ \texttt{move cd 2 to safe 1} ... \newline \textbf{Obs:} Repetition loop detected. & \textbf{Action:} \texttt{take cd 1 from desk 1} $\rightarrow$ \texttt{go to safe 1} $\rightarrow$ \texttt{move cd 1 to safe 1} \\ \midrule
\textbf{Result} & \textbf{\color{red} Failed: Text-Action Disconnect} \newline (Understands the instruction but fails the execution) & \textbf{\color{blue} Success: Task Solved} \newline (Implicitly adheres to the sequential retrieval pattern) \\ \bottomrule
\end{tabularx}
\end{table*}

\end{document}